\documentclass{article}

\PassOptionsToPackage{round}{natbib}
\usepackage[preprint]{neurips_2024}
\usepackage{tikz}
\usepackage{pgfplots}

\usepackage{multirow} 
\usepackage{amsmath}
\usepackage{amssymb}
\usepackage{algorithm}
\usepackage{algorithmic} 
\usepackage{graphicx} 
\usepackage{xcolor} 
\usepackage{tikz} 
\usepackage{forest} 
\usepackage{natbib}

\useforestlibrary{edges}
\colorlet{hidden-draw}{white}

\usepackage[utf8]{inputenc} 
\usepackage[T1]{fontenc}    

\usepackage{hyperref}       
\usepackage{url}            
\usepackage{booktabs}       
\usepackage{amsfonts}       
\usepackage{nicefrac}       
\usepackage{microtype}      
\usepackage{xcolor}         
\usepackage{booktabs}
\usepackage{pifont}
\usepackage{makecell}
\usepackage{ragged2e}
\usepackage{fontawesome5}
\newcommand{\cmark}{\ding{51}} 
\newcommand{\xmark}{\ding{55}} 
\definecolor{physblue}{RGB}{45,105,180}
\definecolor{econorange}{RGB}{190,95,35}

\newcommand{\physkw}[1]{\textcolor{physblue}{\textbf{#1}}}
\newcommand{\econkw}[1]{\textcolor{econorange}{\textbf{#1}}}

\newcommand{\pmark}{%
  \tikz[baseline=-0.55ex, scale=0.09]{
    \draw[line width=0.6pt, gray!70!black] (0,0) circle (1);
    \fill[gray!45] (0,-1) arc[start angle=-90, end angle=90, radius=1] -- cycle;
  }%
}
\usepackage{enumitem}
\usepackage{tabularx}

\usepackage{booktabs}
\usepackage{array}
\usepackage[table]{xcolor}
\usepackage{tikz}
\usetikzlibrary{positioning,calc,arrows.meta}
\usepackage{tcolorbox}
\tcbuselibrary{breakable}

\newcommand{\fullmark}[1]{\textcolor{#1}{\ding{51}}}
\newcommand{\nomark}{\textcolor{black!75}{\ding{55}}}
\newcommand{\partialmark}[1]{%
\tikz[baseline=-0.6ex,scale=0.08]{
  \fill[#1!35] (0,0) -- (0:1) arc[start angle=0,end angle=180,radius=1] -- cycle;
  \draw[line width=0.55pt,#1!80!black] (0,0) circle (1);
}}

\usepackage{comment}
\usepackage{graphicx}

\usepackage{wrapfig}
\usepackage{makecell}
\usepackage{ragged2e}
\usepackage{enumitem}
\usepackage{tabularx}
\usepackage{array}
\usepackage[table]{xcolor}
\usepackage{tikz}
\usetikzlibrary{positioning,calc,arrows.meta}
\usetikzlibrary{shapes.geometric, arrows.meta, positioning, calc, shadows}
\usepackage{pifont}
\definecolor{L1}{RGB}{44, 89, 133}
\definecolor{L2}{RGB}{76, 161, 141}
\definecolor{L3}{RGB}{244, 162, 89}
\definecolor{L4}{RGB}{231, 111, 81}
\definecolor{L5}{RGB}{154, 96, 127}
\definecolor{L6}{RGB}{255, 32, 110}
\newcommand{\levelhead}[3]{%
\textcolor{#1}{\parbox[c][2.55em][c]{\linewidth}{\centering\bfseries\mbox{#2}\\[-1pt]\mbox{\scriptsize #3}}}%
}
\usepackage{amsmath}
\usepackage{amssymb}
\usepackage{algorithm}
\usepackage{algorithmic}
\usepackage{tcolorbox}
\tcbuselibrary{breakable}

\usepackage[normalem]{ulem}
\newtheorem{desideratum}{Desideratum}

\definecolor{EWMRed}{RGB}{180,0,0}
\definecolor{EWMBlue}{RGB}{0,70,160}

\usepackage{listings}
\definecolor{codekeyword}{HTML}{008000}
\definecolor{codecomment}{HTML}{3D7B7B}
\definecolor{codestring}{HTML}{BA2121}
\definecolor{codenumber}{HTML}{666666}
\definecolor{codebuiltin}{HTML}{008000}
\lstdefinestyle{ewmpython}{
  language=Python,
  basicstyle=\ttfamily\footnotesize,
  keywordstyle=\color{codekeyword}\bfseries,
  commentstyle=\color{codecomment}\itshape,
  stringstyle=\color{codestring},
  numberstyle=\color{codenumber}\scriptsize,
  emph={True,False,None,self},
  emphstyle=\color{codebuiltin}\bfseries,
  backgroundcolor=\color{gray!4},
  frame=tb,
  rulecolor=\color{black!35},
  framesep=1mm,
  breaklines=true,
  breakatwhitespace=false,
  columns=fullflexible,
  keepspaces=true,
  showstringspaces=false,
  tabsize=4,
  upquote=true,
  aboveskip=0pt,
  belowskip=0pt
}
\newcommand{\rev}[1]{{\color{black}#1}}

\usepackage{upquote} 
\definecolor{residentblue}{RGB}{0,90,210}
\definecolor{targetorange}{RGB}{225,65,15}

\lstdefinestyle{pythoncode}{language=Python, basicstyle=\ttfamily\footnotesize, keywordstyle=\bfseries\color{blue!60!black}, commentstyle=\itshape\color{gray!70!black}, stringstyle=\color{teal!60!black}, numbers=left, numberstyle=\tiny\color{gray}, stepnumber=1, numbersep=8pt, showstringspaces=false, breaklines=true, breakatwhitespace=true, tabsize=4, columns=fullflexible, keepspaces=true, frame=tb, rulecolor=\color{black!25}, backgroundcolor=\color{gray!4}, captionpos=b, xleftmargin=1.2em, framexleftmargin=1.2em }

\title{From \textit{Economic Agents} to \textit{Agentic Economies}:\\ A Systems Blueprint for Economic World Models
}

\author{
 Jiale Han$^1$, Xiang Li$^{1,2}$, Jing Qian$^{1,2}$, Wenyuan Gu$^2$,
 \textbf{Pin Gao}$^2$, \textbf{Ye Luo}$^{1,3}$, \\\textbf{Hongyuan Zha}$^2$, \textbf{Dacheng Tao}$^4$, \textbf{Benyou Wang}$^{1,2*}$, \textbf{Lin William Cong}$^{1,4}$\thanks{Send correspondence to Cong (\textit{will.cong@ntu.edu.sg}) or Wang (\textit{wangbenyou@cuhk.edu.cn})}\\
  $^1$ Shenzhen Loop Area Institute\\
  $^2$ School of Data Science, The Chinese University of Hong Kong, Shenzhen\\ 
  $^3$ University of Hong Kong  $^4$ Nanyang Technological University \\
  \href{https://economic-world-model.github.io}{%
    \raisebox{-0.04em}{\textcolor{blue!36!white}{\small\faGlobeAmericas}}%
    \hspace{0.24em}%
    \textcolor{black!88}{\nolinkurl{economic-world-model.github.io}}%
  }
}

\begin{document}
\maketitle
\begin{abstract}

Economic World Models (EWMs)~\citep{cong2025ewmddge} are generative economic models that simulate how economies evolve from within by modeling heterogeneous agents, their beliefs and actions, and the market and institutional mechanisms through which their interactions produce aggregate outcomes. 
This paper develops an implementation roadmap for building economic world models as generative engines in which heterogeneous agents act, interact, adapt, and co-evolve with markets and institutions, thereby producing economic dynamics from the inside. We organize EWM systems into a six-level capability ladder, from fixed rule-based agent worlds to adaptive and LLM-based agent worlds, self-evolving agents, evolving institutional worlds, and sim-to-real economic twins aligned with real observations. A systematic literature survey across these levels reveals that existing work remains concentrated in lower-level agent and simulation environments, while systems with self-evolving agents, endogenous institutions, persistent empirical alignment, and validated economic mechanisms remain rare. By translating the EWM agenda into an implementation blueprint, this paper aims to accelerate the development of the next generation of economic simulation environments that can serve as high-fidelity sandboxes for human decision-makers and as training, planning, evaluation, and safety substrates for AI agents. We release a curated paper list and related resources to support future research.\footnote{\url{https://github.com/FreedomIntelligence/Awesome-Economic-World-Models}}

\end{abstract}

\section{Introduction}
\begin{figure}[htb]
  \centering
  \includegraphics[width=\textwidth]{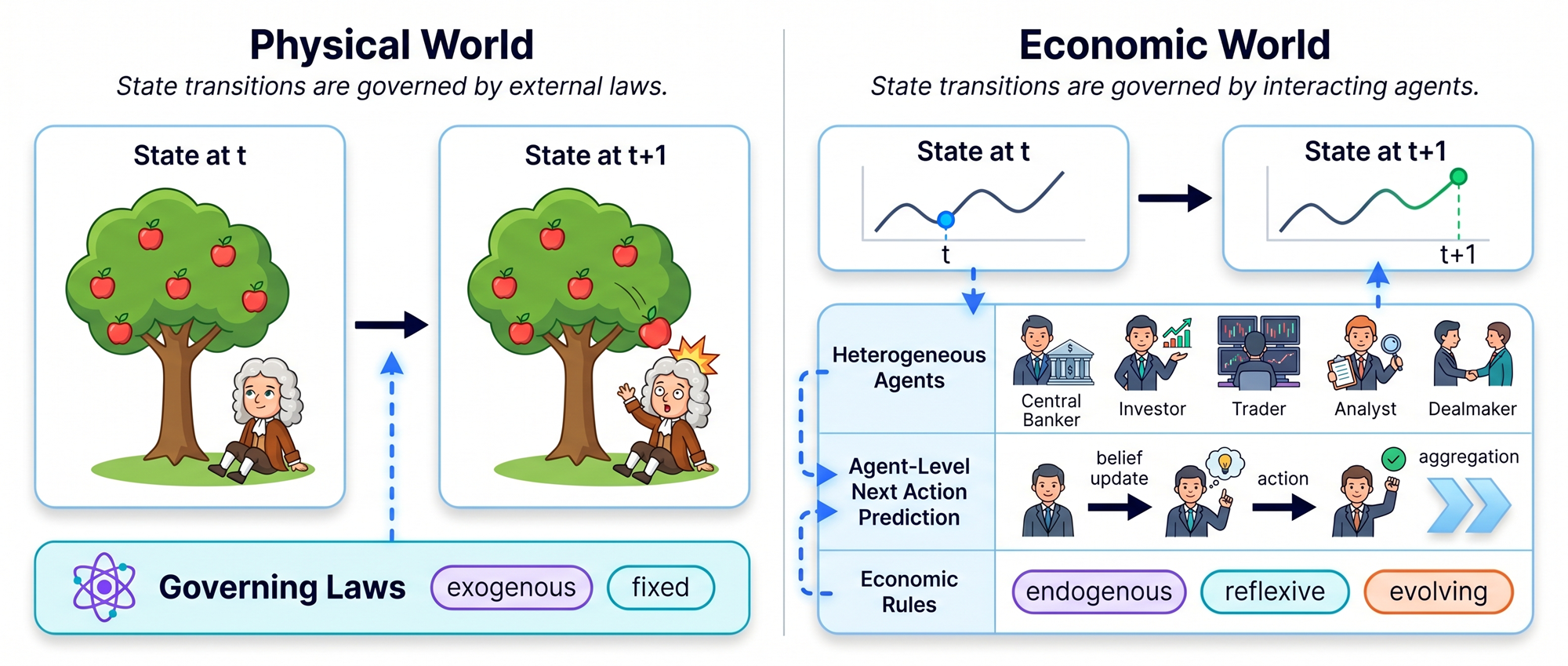}
  \caption{Physical versus economic worlds. Physical state transitions are governed by external laws, whereas economic state transitions are generated by heterogeneous agents interacting through markets, institutions, and evolving rules.}
  \label{fig:intro}
\end{figure}
\begin{quote}
\centering
\emph{``I can calculate the motion of heavenly bodies, but not the madness of people.''}\\
\hfill --- Isaac Newton

\end{quote}

Economics has long advanced through a tradition of ``observe and explain (or predict).'' Economists observe aggregate variables, collect micro-level evidence, estimate empirical relationships or causal effects, build theoretical and quantitative models, and ask what caused what \citep{haavelmo1944probability,friedman1953methodology,heckman2001micro}. This approach, albeit undeniably powerful, leaves open a deeper question of generative explanation. 
Deriving an equilibrium, fitting parameters, or forecasting a turning point does not by itself explain an economic phenomenon unless the outcome can emerge from the model inside. This idea motivates the central premise of this paper: to understand an economy, we should not only observe its outcomes, but build a world simulator which generates such outcomes.


Building on the EWM/DDGE framework of \citet{cong2025ewmddge}~\footnote{This paper is complementary to \citet{cong2025ewmddge}. Cong introduces Economic World Models as an economic model class and develops Data-Driven Generative Equilibrium (DDGE) as the equilibrium discipline needed when behavior, beliefs, data generation, and learned environments are jointly endogenous. The present paper takes a CS/AI systems perspective: it asks how such economic worlds can be implemented, evaluated, and scaled as computational environments for agent training, policy sandboxes, planning engines, and sim-to-real economic twins. Accordingly, the capability levels below classify implementation maturity; they do not by themselves guarantee DDGE-style counterfactual consistency.}, we use the following AI-systems working definition for implementation purposes:
\begin{quote} An  Economic World Model system is a \textbf{generative engine for economic environments}: it predicts how an economy moves by modeling how the agents inside observe, reason, act, interact, and adapt under economic institutions and constraints. \end{quote}

\textbf{Economic vs. Physical World Model.}~
This perspective connects EWMs to the broader idea of world models in AI \citep{ha2018world,ding2025understanding}. 
A world model learns the dynamics of an environment so that an agent can imagine possible futures, evaluate actions before taking them, and plan in simulation. 
In this sense, an Economic World Model extends the general idea of world modeling from physical environments to economic environments.
Newton's remark captures precisely why economic worlds are different: 
{\tt the motion of planets can be described by external laws that do not depend on what planets believe}. 
The key difference lies in what drives the transition from one state to the next. 
In physical worlds, state transitions are primarily governed by stable external laws \footnote{For example, objects move under forces that do not depend on their own intentions or beliefs.}, as shown in Figure~\ref{fig:intro}.
In economic worlds, by contrast, transitions are generated endogenously by agents themselves. 
Households, firms, banks, and regulators observe information, form expectations, make decisions, and interact through markets and institutions. 
Their collective behavior produces the next economic state, which in turn reshapes their future beliefs and actions.

\paragraph{Link between agents and EWMs.}
Agents and the Economic World Model are, in some sense, two sides of the same coin, with the former taking a microeconomic perspective and the latter a macroeconomic one.
\begin{itemize} 
    \item An (economic) \textbf{agent} learns what an individual economic entity will do. Given the current economic state $s_t$, they observe information, form beliefs, face constraints, and choose economically feasible actions $a_t = \mathcal{\pi}(s_t)$.
    \item The \textbf{Economic World Model} predicts what the economy will become. Given the current economic state $s_t$ and interventions $u_t$, it maps them to the next economic state, $s_{t+1} = \mathcal{T}(s_t, u_t)$.
\end{itemize}
This action--state loop forms the core runtime of an EWM. Agents submit actions, and the world model transforms them into a next economic state. This updated state then shapes agents' subsequent beliefs, constraints, memories, and strategies. Richer worlds expose agents to scarcity, incentives, institutions, and strategic feedback. In turn, more realistic agents generate adaptive responses, edge cases, and failure modes that make the world model more faithful.

\begin{figure}[H]
  \centering
  \includegraphics[width=0.9\linewidth]{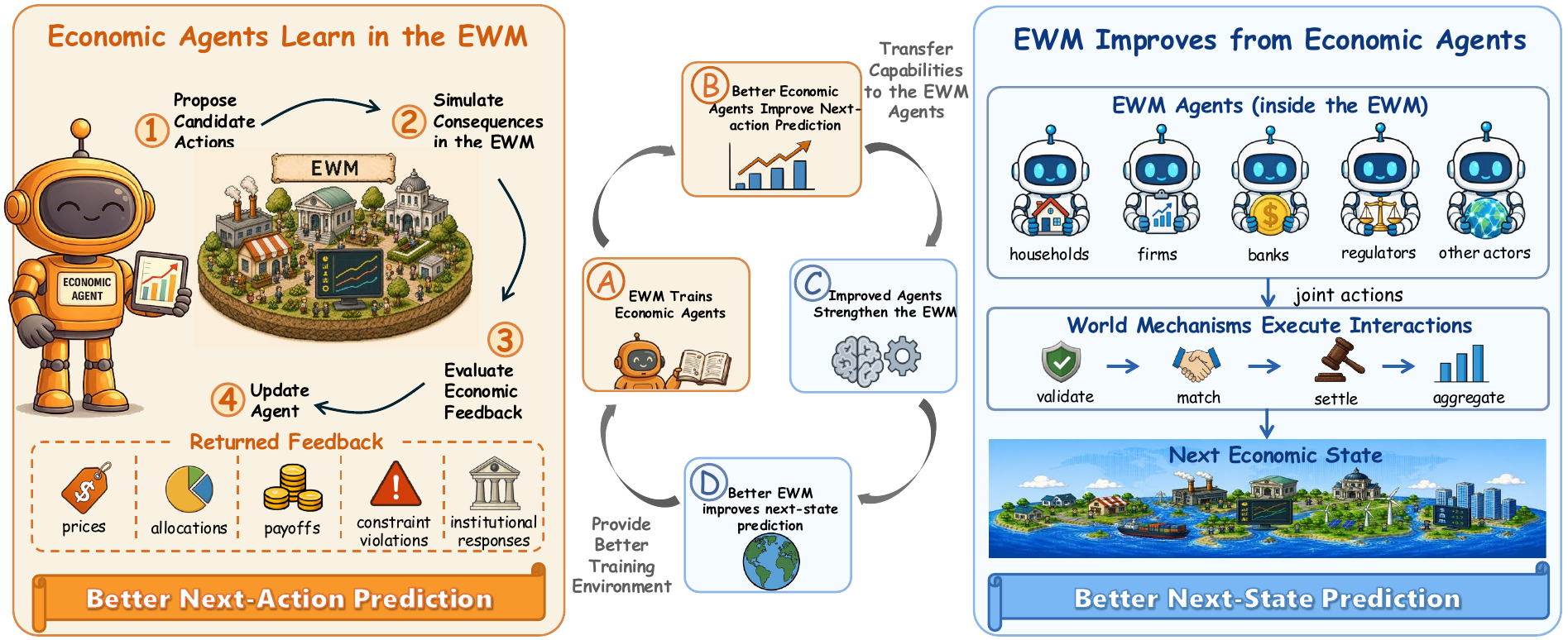}
  \caption{Iterative co-improvement between Economic Agents and an EWM.}
  \label{fig:agentic-economy-improvement}
\end{figure}

\textbf{The goal: towards agentic economies.}~
The goal is to move from isolated \emph{Economic Agents} to full \emph{Agentic Economies}.
A stand-alone \emph{Economic Agent} primarily performs agent-side next-action generation: given an observed economic state, it predicts or selects the next feasible action of an individual economic actor.
What it lacks is an executable economic feedback loop in which the consequences of that action are realized, evaluated, and returned as learning signals. An \emph{Agentic Economy}, by contrast, turns this one-sided prediction problem into an engineered closed loop between agent-side next-action prediction and world-side next-state prediction. Before acting in the real economy, an agent can first interact with an EWM through simulated rollouts to anticipate consequences, evaluate alternative actions, and choose a better-informed action. As illustrated in Figure~\ref{fig:agentic-economy-improvement}, the EWM and economic agents can improve each other through an iterative training loop. Agents can be trained through simulated rollouts in the EWM.
As these agents become more capable or behaviorally accurate, these advances can be transferred to the agents in the EWM.
This improves next action prediction of  EWM agents, thereby leading to more accurate next-state prediction.
The resulting EWM then provides a more realistic and informative environment for subsequent agent training.
Repeating this process creates a co-improvement cycle between economic agents and the EWM.

\paragraph{Implementation.}
Operationally, we implement an EWM as a modular economic runtime rather than a monolithic predictor. 
The runtime couples four layers: an \emph{agent layer} that instantiates heterogeneous economic actors with objectives, information, constraints, memory, tools, and feasible actions; an \emph{environment layer} that encodes economic states, market mechanisms, contracts, accounting identities, and institutional rules; a \emph{co-evolution layer} that lets agents, strategies, mechanisms, and institutions adapt over repeated rollouts; and a \emph{real-world alignment layer} that compares simulated trajectories with empirical observations and corrects model drift. 
At each step, agents observe the current state and choose actions, constraint validators check feasibility, mechanism engines execute and aggregate interactions, the world publishes the next economic state, and agents update their beliefs and strategies. 
This runtime view makes EWMs buildable: agents generate actions, the world model generates states, and empirical alignment keeps the artificial economy connected to reality.

\paragraph{Why now?}
Recently, frontier AI labs are beginning to institutionalize economic expertise as part of AI development. 
OpenAI appointed its first Chief Economist in 2024 \citep{openai2024chatterji} and subsequently launched the OpenAI Economic Research Exchange in 2026 \citep{openai2026exchange}. Anthropic has created the Anthropic Economic Index to study AI's effects on labor markets, firms, institutions, and the broader economy \citep{anthropic2025economicindex}. These highlight the growing need to understand AI's economic consequences. 
Technically, large language models provide a substrate for modeling beliefs, memory, communication, reasoning, and context-sensitive decision-making \citep{an2024make,xu2026mem,ferrag2025llm,guo2025deepseek,zhu2024language}. Tool-using agents \citep{masterman2024landscape,shi2025tool} connect decisions to executable economic actions such as search, purchase, negotiation, trading, planning, and compliance. The rise of agentic AI further turns agents into actors that can plan, call tools, invoke reusable skills, and execute multi-step tasks through external harnesses \citep{zhou2026externalization}. Multi-agent simulation \citep{chen2024survey} provides an interaction substrate in which heterogeneous agents can respond to one another and jointly generate aggregate outcomes. Meanwhile, the growing availability of financial, textual, policy, administrative, and transaction data provides the empirical foundation for constructing, calibrating, and validating EWMs. Taken together, these developments bring Economic World Models within reach.

\paragraph{Applications of EWMs.}
With an EWM system, economic reasoning becomes operational and experimental. For human decision-makers, the system functions as a high-fidelity sandbox for testing policies, market designs, firm strategies, systemic risks, and institutional reforms before deployment. For AI systems, it functions as a training, planning, and evaluation substrate: an environment where agents learn under scarcity, incentives, and institutional constraints; a simulator for testing downstream consequences before action; and a safety testbed for detecting interaction risks such as manipulation, collusion, instability, and cascading failure.



\paragraph{Position.}
Finance already provides several component-level prototypes for this systems agenda. 
\citet{cong2021alphaportfolio} illustrate an agent-side decision module, training a 
portfolio agent directly on economic objectives through deep reinforcement learning. 
\citet{campello2025alphamanager} illustrate a learned environment module and robust 
decision layer for corporate policy search. \citet{cong2024writing} illustrate an 
agent--data--model feedback module in online credit, where borrower-side generated text 
changes the data observed by lender-side models. \citet{bini2025behavioral} illustrate 
agent-evaluation and behavioral-calibration modules for LLM decision agents. These 
studies are not complete EWM systems under the taxonomy below; they are reusable 
modules for building such systems.
This paper translates the EWM agenda into a CS/AI systems roadmap by articulating the implementation vision, operationalizing core concepts, building a capability ladder, and providing an architectural starting point. 
Building trustworthy, scalable, and empirically grounded EWMs will require a community effort across AI, economics, management, social science, and policy. 
Our goal is to make economic worlds buildable, testable, and alignable as AI systems, while preserving the economic discipline of the EWM/DDGE framework.


\section{Economic World Models as AI Systems}
\label{sec:definition}

\subsection{An Implementation-Oriented Working Definition}

Generally, a \textbf{world model} is a computable dynamical system that represents how the state of a world evolves over time, especially under actions or interventions. Its core role is to approximate the transition dynamics of an environment: given a current state and an action, it predicts the next state. Formally, a world model can be written as
\begin{equation}
\hat{s}_{t+1} = \mathcal{T}\bigl(s_t, u_t\bigr),
\end{equation}
where \(s_t\) is the current state, \(u_t\) is an external intervention or exogenous shock, $\hat{s}_{t+1}$ is the predicted next state, and \(\mathcal{T}\) denotes the transition operator. 

\paragraph{Physical Worlds vs.\ Economic Worlds.}

\begin{wraptable}[14]{r}{0.52\linewidth}
\vspace{-1em}
\centering
\caption{Physical vs. economic worlds.}
\label{tab:physical-vs-economic}
\scriptsize
\setlength{\tabcolsep}{3pt}
\renewcommand{\arraystretch}{1.18}
\begin{tabular}{@{}p{0.25\linewidth}p{0.32\linewidth}p{0.35\linewidth}@{}}
\toprule
\textbf{Dimension} 
& \textcolor{physblue}{\textbf{Physical}} 
& \textcolor{econorange}{\textbf{Economic}} \\
\midrule
Main driver      
& \physkw{Laws}\newline {\footnotesize external dynamics}
& \econkw{Beliefs \& incentives}\newline {\footnotesize subjective response} \\

Actors           
& \physkw{Objects}\newline {\footnotesize passive entities}
& \econkw{Strategic agents}\newline {\footnotesize decision makers} \\

Outcomes         
& \physkw{Externally governed}\newline {\footnotesize law-implied results}
& \econkw{Endogenously formed}\newline {\footnotesize interaction outcomes} \\

Feedback         
& \physkw{Mechanical}\newline {\footnotesize physical causality}
& \econkw{Reflexive}\newline {\footnotesize beliefs reshape actions} \\

Rules            
& \physkw{Fixed}\newline {\footnotesize stable laws}
& \econkw{Evolving}\newline {\footnotesize institutions adapt} \\

State transition 
& \physkw{Law-based}\newline {\footnotesize object dynamics}
& \econkw{Interaction-based}\newline {\footnotesize agents generate states} \\
\bottomrule
\end{tabular}
\vspace{-1.2em}
\end{wraptable}

The core distinction between physical worlds and economic worlds is the distinction between \textbf{objectivity} and \textbf{subjectivity}. Physical worlds are primarily objective transition systems: objects move according to external laws that do not depend on what the objects believe, expect, or intend. Economic worlds are shaped by subjective agents who interpret information, form beliefs, make strategic decisions, and revise their behavior over time. Table~\ref{tab:physical-vs-economic} summarizes the core contrast. Four features make economic worlds especially distinct.

\begin{itemize}[leftmargin=*]
    \item \textbf{Subjective agents:} Unlike physical objects, economic agents interpret the world before acting. Households, firms, banks, investors, and regulators form beliefs, expectations, and strategic views under incomplete information and institutional constraints.
    \item \textbf{Endogenous outcomes:} Core economic variables such as prices, allocations, quantities, risks, and macroeconomic conditions are not externally imposed. They emerge from decentralized interactions such as trading, bargaining, bidding, contracting, and market clearing.
    \item \textbf{Reflexive feedback:} Economic agents act on their beliefs, and their actions change the economy. The changed economy then reshapes later beliefs and decisions. Economic feedback is therefore not merely mechanical, but belief-mediated and self-referential.
    \item \textbf{Adaptive dynamics:} Economic worlds do not evolve under fixed behavioral laws. Agents learn, strategies change, expectations shift, and institutions or policy rules may adjust over time. The transition dynamics of the economy therefore co-evolve with the agents inside it.
\end{itemize}


\paragraph{State in Economic Worlds.}
For CS/AI implementation, we use EWM to denote a computable dynamical system that captures the transition of an economy driven by heterogeneous and interacting agents. This working definition is narrower and more operational than the economic framework in \citet{cong2025ewmddge}: it focuses on system components, capability levels, and implementation requirements rather than on equilibrium concepts.

A central implementation question is what counts as the ``state'' of an economic world. Unlike a physical state, which can often be described by objective variables of objects, an economic state must contain both objective conditions and subjective agent-side representations. It is therefore not merely a vector of observable market variables, but a typed snapshot that makes the next economic transition executable. We write
\begin{equation}
s_t = \bigl( x_t,\; \{z_t^i\}_{i=1}^{N_t},\; \{b_t^i\}_{i=1}^{N_t},\; \mathcal{I}_t \bigr),
\end{equation}
where the state contains three parts:
\begin{itemize}
    \item \textbf{Aggregate economic state \(x_t\).} This includes economic variables and information such as prices, quantities, volatility, employment, output, interest rates, public news and other economy-wide conditions.

    \item \textbf{Agent-side states \(\{z_t^i, b_t^i\}_{i=1}^{N_t}\).} These describe both the objective and subjective conditions of each agent. The private state \(z_t^i\) includes cash, wealth, inventory, balance-sheet position, contracts, production capacity, risk exposure, or outstanding obligations. The belief state \(b_t^i\) captures expectations, perceived risks, forecasts, uncertainty, and beliefs about prices, demand, policies, or other agents' behavior.

    \item \textbf{Institutional environment \(\mathcal{I}_t\).} This specifies the rules under which agents act, including constraints, market mechanisms, policy regimes, disclosure rules, information channels, and data-access permissions.
\end{itemize}

\paragraph{\textbf{State transition in an EWM.}}
Given an intervention \(u_t\), the transition operator decomposes into the following stages, as summarized in Figure~\ref{fig:ewm_transition}:
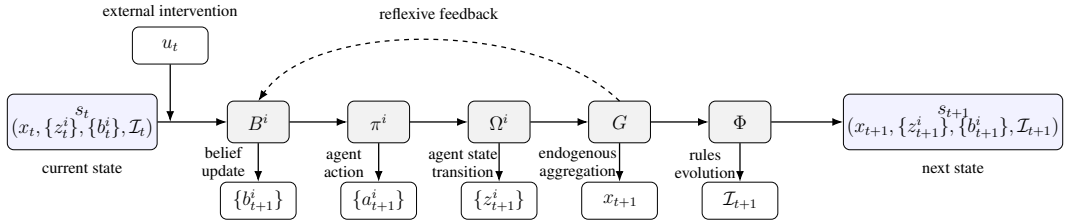
\begin{figure}[h]
\centering
\resizebox{1\columnwidth}{!}{%
\begin{tikzpicture}[>=Latex, font=\large]

\tikzset{
stateL/.style={draw, rounded corners, align=center, minimum height=12mm, text width=30mm, fill=blue!5},
stateR/.style={draw, rounded corners, align=center, minimum height=12mm, text width=45mm, fill=blue!5},
op/.style={draw, rounded corners, align=center, minimum height=9mm, text width=11mm, fill=gray!10},
var/.style={draw, rounded corners, align=center, minimum height=8mm, text width=14mm},
lab/.style={font=\normalsize, text=black, align=center}
}

\node[stateL] (st)  at (0,0) {$s_t$\\[-1mm]$(x_t,\{z_t^i\},\{b_t^i\},\mathcal{I}_t)$};
\node[op]     (B)   at (3.8,0) {$B^i$};
\node[op]     (Pi)  at (6.4,0) {$\pi^i$};
\node[op]     (Om)  at (9.0,0) {$\Omega^i$};
\node[op]     (G)   at (11.6,0) {$G$};
\node[op]     (Phi) at (14.2,0) {$\Phi$};
\node[stateR] (st1) at (18.8,0) {$s_{t+1}$\\[-1mm]$(x_{t+1},\{z_{t+1}^i\},\{b_{t+1}^i\},\mathcal{I}_{t+1})$};

\draw[->, thick] (st) -- (B);
\draw[->, thick] (B) -- (Pi);
\draw[->, thick] (Pi) -- (Om);
\draw[->, thick] (Om) -- (G);
\draw[->, thick] (G) -- (Phi);
\draw[->, thick] (Phi) -- (st1);

\node[lab] at (0,-1.0) {current state};
\node[lab] at (18.8,-1.0) {next state};

\node[var] (actbox) at (1.9,1.65) {$u_t$};
\node[lab] at (1.9,2.4) {external intervention};
\draw[->, thick] (actbox.south) -- (1.9,0);

\node[var] (bt1) at (3.8,-1.7) {$\{b_{t+1}^i\}$};
\node[var] (at1) at (6.4,-1.7) {$\{a_{t+1}^i\}$};
\node[var] (zt1) at (9.0,-1.7) {$\{z_{t+1}^i\}$};
\node[var] (xt1) at (11.6,-1.7) {$x_{t+1}$};
\node[var] (It1) at (14.2,-1.7) {$\mathcal{I}_{t+1}$};

\draw[->, thick] (B) -- (bt1);
\draw[->, thick] (Pi) -- (at1);
\draw[->, thick] (Om) -- (zt1);
\draw[->, thick] (G) -- (xt1);
\draw[->, thick] (Phi) -- (It1);

\node[lab] at (3.05,-0.9) {belief\\update};
\node[lab] at (5.65,-0.9) {agent\\action};
\node[lab] at (8.22,-0.9) {agent state\\transition};
\node[lab] at (10.7,-0.9) {endogenous\\aggregation};
\node[lab] at (13.45,-0.9) {rules \\evolution};

\draw[->, thick, dashed] (11.6,0.45) .. controls (9.8,2.2) and (5.0,2.2) .. (3.8,0.45);
\node[lab] at (7.7,2.35) {reflexive feedback};
\end{tikzpicture}%
}
\caption{State transition in an Economic World Model.}
\label{fig:ewm_transition}
\end{figure}

{\small\begin{align}
b_{t+1}^i 
&= B^i\bigl(b_t^i,\; z_t^i,\; x_t,\; u_t,\; \mathcal{I}_t\bigr),
\quad i=1,\ldots,N_t
&& \text{(agent belief update)}\\
a_{t+1}^i 
&= \pi^i\bigl(z_t^i,\; b_{t+1}^i,\; x_t,\; u_t,\; \mathcal{I}_t\bigr),
\quad i=1,\ldots,N_t
&& \text{(\textbf{next-action prediction of agents})}\\
z_{t+1}^i 
&= \Omega^i\bigl(z_t^i,\; a_{t+1}^i\bigr),
\quad i=1,\ldots,N_t
&& \text{(agent state update)}\\
x_{t+1} 
&= G\bigl(\{z_{t+1}^i,\; a_{t+1}^i,\; b_{t+1}^i\}_{i=1}^{N_t},\; x_t,\; \mathcal{I}_t\bigr)
&& \text{(\textbf{next-state prediction of economic worlds})}\\
\mathcal{I}_{t+1} 
&= \Phi\!\left(\mathcal{I}_t,\; x_{t+1},\;
\{b_{t+1}^i,\; a_{t+1}^i,\; z_{t+1}^i\}_{i=1}^{N_t}\right)
&& \text{(rules and mechanisms evolution)}
\end{align}}Where $B^i$ denotes the belief-update operator of agent $i$, mapping the agent's prior belief, current state, aggregate economic state, exogenous intervention, and institutional environment into an updated subjective representation of the economy. Based on this, $\pi^i$ represents the decision rule mapping the agent's current state and updated belief to an action, while $\Omega^i$ is the state transition function updating individual micro-level variables such as wealth, inventory, or balance-sheet positions. Finally, $G$ defines the endogenous aggregation and market-closure operator that maps the collection of individual agent states into next-period aggregate outcomes (e.g., prices, allocations, and macroeconomic conditions), and $\Phi$ governs the evolution of institutions or governing mechanisms. 


\rev{\subsection{An Execution Interface for EWM Systems}
\label{subsec:execution-interface}
The equations above describe an EWM at the level of state-transition operators. Figure~\ref{fig:ewm-interface} gives this formal object a minimal execution interface. The code follows the pattern of an environment package: \texttt{ewm.make} constructs a named economic world with heterogeneous agent populations, \texttt{world.reset} initializes the first economic state, \texttt{world.run\_agents(state, parallel=True)} collects actions from all agents in parallel, \texttt{world.step(actions)} advances the economy with the joint action dictionary, and \texttt{world.close} releases runtime resources. This usage-first interface keeps the public loop compact while allowing richer mechanisms, constraints, logging, evolution, and alignment to live behind the runtime.}

\begin{figure}[ht]
\centering
\begin{minipage}{0.9\textwidth}
\begin{lstlisting}[style=ewmpython]
import economic_world_model as ewm

# Initialise the economic world
world = ewm.make(
     name="FXMarket-v0",
     agents=[ewm.agent(role="household", count=1000),
             ewm.agent(role="firm", count=20),
             ewm.agent(role="bank", count=5)],
     environment = ewm.environment("FXMarket-v0"))

# Reset the world to generate the first economic state
state = world.reset(seed=42)

for t in range(T):
    # this is where agents observe the state and choose economic actions
    actions = world.run_agents(state, parallel=True) 

    # step through the economic world with joint actions
    # receiving the next state
    state = world.step(actions)

world.close()
\end{lstlisting}
\end{minipage}
\caption{A minimal execution interface for EWM systems. }
\label{fig:ewm-interface}
\end{figure}

\rev{The example should be read as an environment-style rollout for an artificial economy. The identifier \texttt{"FXMarket-v0"} selects a reusable economic scenario, while the \texttt{agents} argument specifies the population through role-level declarations for households, firms, and banks. After \texttt{world.reset(seed=42)}, the returned \texttt{state} contains the current economic condition and runtime context such as observations, diagnostics, and seed-dependent metadata. In each iteration, \texttt{world.run\_agents(state, parallel=True)} concurrently elicits decisions from all agents, returning an \texttt{actions} dictionary that maps each agent’s identifier to its chosen economic action. The call \texttt{world.step(actions)} returns the updated \texttt{state}. The minimal code omits optional reset-on-termination logic and advanced hooks such as logging, evolution, and real-data alignment. Section~\ref{sec:reusable_implementation} develops those API contracts behind this compact loop. 
}

\subsection{Engineering Desiderata for EWM Systems}

For implementation, a strong EWM system should satisfy four engineering desiderata:


\begin{desideratum}
\textbf{Endogenous closure.} {An EWM system should produce key economic outcomes endogenously.}
\end{desideratum}
Prices, allocations, and macroeconomic conditions should emerge from decentralized interactions among heterogeneous agents, rather than be imposed from outside. This matters because these outcomes do not simply describe the state of the economy. They also shape what agents do next by feeding back into subsequent decisions and interactions. If they are treated as exogenous, the model can no longer represent the feedback process through which an economic world evolves.

\begin{desideratum}
\textbf{Behavioral fidelity.} {Agents in an EWM system should resemble real economic actors.}
\end{desideratum}
Agent behavior should reflect how households, firms, banks, and investors perceive, interpret, and respond to incentives, information, and constraints. Such fidelity may be realized in different ways, ranging from symbolic, non-LLM cognitive substrates that support in-world strategy adaptation in response to economic signals, to richer LLM-based cognitive substrates that maintain explicit beliefs, expectations, and subjective representations of the economy. What matters is capturing heterogeneity, bounded rationality, and context-dependent adaptation, rather than reducing behavior to fixed or overly stylized rules.
Behavioral fidelity is especially important when LLMs are used as economic agents.
Recent evidence shows that LLMs can exhibit systematic behavioral patterns in 
economic and financial decision tasks, including biases in preference-based tasks and 
more rational responses in some belief-based tasks; prompting and correction 
procedures can mitigate some of these biases \citep{bini2025behavioral}. EWM 
agents therefore require behavioral validation rather than persona prompts alone.

\begin{desideratum}
\textbf{Evolving dynamics.} {In an EWM system, both the agents and the governing rules of the economy should co-evolve over time.}
\end{desideratum}
Economic worlds are not static: strategies adapt, beliefs shift, agents may even acquire new cognitive capabilities and behavioral repertoires through experience, and institutional conditions and governing rules evolve endogenously. A credible EWM should therefore model not only state transitions, but also the evolution of strategies, cognition, and governing rules at each of these layers.

\begin{desideratum}
\textbf{Reality alignment.} {An EWM system should stay aligned with the real economy over time.}
\end{desideratum}
It should not only generate plausible simulations, but also repeatedly compare them with newly observed outcomes and update its internal dynamics accordingly. In this way, through repeated sim-to-real alignment, the model can progressively narrow the gap between the simulated economy and the real one.

\section{A Systematic Empirical Study of the EWM Literature Landscape}
\label{sec:toward}

The previous section defines EWM systems as AI systems that generate economic state transitions through heterogeneous agents, economic mechanisms, and institutional constraints. We now study how this vision has been implemented in existing research. 

\subsection{Six Capability Levels of EWM Systems}
\label{subsec:capability-framework}
We organize EWM systems into six progressively stronger capability levels. The levels translate the engineering desiderata in Section~\ref{sec:definition} into observable implementation criteria. 

\begin{table}[ht!]
\centering
\caption[Capability levels of EWM systems as progressively stronger realizations of the engineering desiderata.]{
Capability levels of EWM systems as progressively stronger realizations of the engineering desiderata.
Symbols indicate full, partial, and missing satisfaction: \protect\fullmark{L1}, \protect\partialmark{L1}, and \protect\nomark. Agent capability improves from L1 to L4, rules remain fixed through L4 and evolve at L5, and real-time correction appears at L6.}
\label{tab:levels-desiderata}
\scriptsize
\setlength{\tabcolsep}{2.2pt}
\renewcommand{\arraystretch}{1.00}
\begin{tabularx}{0.95\textwidth}{@{}>{\RaggedRight\arraybackslash}p{2.45cm}*{6}{>{\centering\arraybackslash}X}@{}}
\toprule
\rowcolor{gray!8}
\parbox[c][2.55em][c]{\linewidth}{\textbf{Desideratum}} &
\levelhead{L1}{L1 Fixed-rule}{agent worlds} &
\levelhead{L2}{L2 Adaptive}{agent worlds} &
\levelhead{L3}{L3 LLM-based}{agent worlds} &
\levelhead{L4}{L4 Self-evol.}{agent worlds} &
\levelhead{L5}{L5 Evolving}{econ worlds} &
\levelhead{L6}{L6 Sim-to-real}{econ twins} \\
\midrule
\mbox{\textbf{Endogenous closure}}
& \fullmark{L1} & \fullmark{L2} & \fullmark{L3} & \fullmark{L4} & \fullmark{L5} & \fullmark{L6} \\
\mbox{\textbf{Behavioral fidelity}}
& \partialmark{L1} & \partialmark{L2} & \fullmark{L3} & \fullmark{L4} & \fullmark{L5} & \fullmark{L6} \\
\mbox{\textbf{Evolving dynamics}}
& \nomark & \partialmark{L2} & \partialmark{L3} & \partialmark{L4} & \fullmark{L5} & \fullmark{L6} \\
\mbox{\textbf{Reality alignment}}
& \nomark & \nomark & \nomark & \nomark & \nomark & \fullmark{L6} \\
\bottomrule
\end{tabularx}

\vspace{0.10em}
\begin{tikzpicture}[
    trim left=0pt, trim right=0.95\textwidth,
    x=1pt, y=1cm,
    axislabel/.style={font=\bfseries\scriptsize, anchor=west},
    note/.style={font=\fontsize{6.6}{7.1}\selectfont, align=center},
    guide/.style={black!12, line width=0.3pt},
    dashedaxis/.style={dash pattern=on 2.1pt off 2.1pt, line width=0.65pt},
    axisarrow/.style={-{Latex[length=2mm,width=1.35mm]}, line width=0.85pt}
]
\pgfmathsetlengthmacro{\totalW}{0.95\textwidth}
\pgfmathsetlengthmacro{\colOneW}{2.45cm}
\pgfmathsetlengthmacro{\tabsep}{2.2pt}
\pgfmathsetlengthmacro{\stepW}{(\totalW - \colOneW)/6}

\pgfmathsetmacro{\cOne}{(\colOneW + \tabsep + 0.5*\stepW)/1pt}
\pgfmathsetmacro{\cTwo}{\cOne + \stepW/1pt}
\pgfmathsetmacro{\cThree}{\cOne + 2*\stepW/1pt}
\pgfmathsetmacro{\cFour}{\cOne + 3*\stepW/1pt}
\pgfmathsetmacro{\cFive}{\cOne + 4*\stepW/1pt}
\pgfmathsetmacro{\cSix}{\cOne + 5*\stepW/1pt}

\pgfmathsetmacro{\midOneFour}{(\cOne+\cFour)/2}
\pgfmathsetmacro{\midOneFive}{(\cOne+\cFive)/2}

\def\xLabel{0}
\def\yAgent{0}
\def\yRules{-0.46}
\def\yReal{-0.92}

\foreach \x in {\cOne,\cTwo,\cThree,\cFour,\cFive,\cSix} {
  \draw[guide] (\x,0.18) -- (\x,-1.14);
}

\node[axislabel] at (\xLabel,\yAgent) {Agent};
\draw[axisarrow, color=L4] (\cOne,\yAgent) -- (\cFour,\yAgent);
\draw[dashedaxis, color=L4] (\cFour,\yAgent) -- (\cSix,\yAgent);
\node[note, above=2pt, text=black!70] at (\cOne,\yAgent) {fixed};
\node[note, above=2pt, text=L2] at (\cTwo,\yAgent) {adaptivity};
\node[note, above=2pt, text=L3] at (\cThree,\yAgent) {autonomy};
\node[note, above=2pt, text=L4] at (\cFour,\yAgent) {self-evolution};

\node[axislabel] at (\xLabel,\yRules) {Rules};
\draw[dashedaxis, color=black!38] (\cOne,\yRules) -- (\cFour,\yRules);
\draw[axisarrow, color=L5] (\cFour,\yRules) -- (\cFive,\yRules);
\draw[dashedaxis, color=L5!55] (\cFive,\yRules) -- (\cSix,\yRules);
\node[note, above=2pt, text=black!70] at (\midOneFour,\yRules) {fixed rules};
\node[note, above=2pt, text=L5] at (\cFive,\yRules) {rules evolve};

\node[axislabel] at (\xLabel,\yReal) {Real world};
\draw[dashedaxis, color=black!28] (\cOne,\yReal) -- (\cFive,\yReal);
\draw[axisarrow, color=L6] (\cFive,\yReal) -- (\cSix,\yReal);
\node[note, above=2pt, text=black!60] at (\midOneFive,\yReal) {no online correction};
\node[note, above=2pt, text=L6] at (\cSix,\yReal) {online correction};
\end{tikzpicture}
\end{table}

As shown in Table~\ref{tab:levels-desiderata}, EWM systems can be organized into six capability levels that operationalize the four engineering desiderata. 
The hierarchy is an implementation taxonomy rather than an equilibrium concept: a system can be high on the capability ladder and still fail counterfactual consistency if learned components are not re-centered on the data generated by the counterfactual itself, the problem addressed by DDGE in \citet{cong2025ewmddge}.

\noindent\textbf{Level 1: EWM with fixed rule-based agents.} \textit{This level models the economy as a closed system in which \textbf{agents and the environment follow fixed, pre-specified rules}.} At L1, the economy is modeled with fixed rule-based agents. The agent decision rule \(\pi^i\), the agent state transition \(\Omega^i\), and the aggregation operator \(G\) are specified ex ante. The cognitive operator \(B^i\) is inactive, and the institutional evolution operator \(\Phi\) is also inactive. Agents interact under fixed rules, and their actions generate aggregate outcomes. This level provides endogenous closure, but it has no in-world adaptation.

\noindent\textbf{Level 2: EWM with adaptive rule-based agents.} 
\textit{This level moves \textbf{from fixed to adaptive rule-based agents}, where agents revise their decision policies through symbolic or algorithmic mechanisms.} 
At L2, agents are no longer fixed rule executors. The decision rule becomes time-varying, written as \(\pi^i_t\), and is updated using the agent's interaction history and realized aggregate outcomes. The adaptation substrate remains symbolic or algorithmic, such as reinforcement learning, evolutionary search, online learning, or adaptive heuristics. The cognitive operator \(B^i\) is still inactive, so agents do not form rich beliefs or expectations. The institutional evolution operator \(\Phi\) also remains inactive. This level adds in-world adaptation, but the adaptation is still rule-based rather than cognition-based.

\noindent\textbf{Level 3: EWM with LLM-based autonomous agents.} 
\textit{This level moves \textbf{from adaptive rules to LLM-based autonomous agents}, where agents reason with beliefs, memory, and language.} 
At L3, agents are equipped with explicit cognitive states, including beliefs, expectations, memory, and subjective representations of the economy. The cognitive operator \(B^i\) becomes active and updates these states over time. The decision rule \(\pi^i_t\) is conditioned on updated beliefs, agent states, aggregate conditions, and institutional constraints. This allows agents to reason, communicate, and make context-sensitive economic decisions. The cognitive substrate itself remains fixed during the rollout, and \(\Phi\) remains inactive. This level improves behavioral fidelity, but agents do not yet acquire new capabilities.

\noindent\textbf{Level 4: EWM with evolving agents.} \textit{This level moves beyond \textbf{belief-driven autonomy} to \textbf{agent self-evolution}, where agents persistently expand their strategies, skills, or behavioral routines.}
At L4, adaptation extends from beliefs and actions to the agent capability substrate itself. Agents may acquire new strategies, skills, tools, memories, or behavioral routines through in-world experience. These acquired capabilities persist and reshape \(B^i\) and \(\pi^i_t\) in later periods. The update is driven by simulated feedback such as payoffs, interaction outcomes, reflection, and aggregate economic states. The institutional environment remains fixed, so \(\Phi\) is still inactive. This level makes agents cognitively dynamic, but the world around them does not yet evolve.

\noindent\textbf{Level 5: EWM with agent--environment co-evolution.} 
\textit{This level moves \textbf{from evolving agents to agent--environment co-evolution}, where rules and institutions evolve endogenously.} 
At L5, adaptation extends beyond agents to the governing structure of the economy. The institutional evolution operator \(\Phi\) becomes active, allowing policies, market rules, contracts, governance structures, or institutional constraints to change in response to agent behavior and aggregate outcomes. The agent substrate may correspond to L2, L3, or L4. What defines this level is not only how intelligent agents are, but whether the rules of the game become endogenous. This level creates an evolving economic world, but it is not yet repeatedly corrected by real-world observations.

\noindent\textbf{Level 6: sim-to-real economic twins.} 
\textit{This level moves \textbf{from internally evolving worlds to online sim-to-real alignment}, where the running model is corrected as new real-world observations arrive.} At L6, the EWM is no longer a closed simulator that evolves only through internal dynamics. It maintains an online correction loop that compares simulated trajectories with newly observed economic data during the rollout. When discrepancies arise, the system updates relevant components, including agent behavior, parameters, mechanisms, institutional rules, or state-transition modules. This creates a two-loop architecture. The inner loop generates endogenous economic dynamics, while the outer loop performs real-time monitoring, diagnosis, and correction. This level adds reality alignment and moves the system toward a self-correcting digital twin of the economy.

\subsection{Literature Collection and Classification}
\label{subsec}

We systematically survey the existing literature and classify each
validated paper using the six-level coding rules in Table~\ref{tab:levels-desiderata}.

\tikzstyle{my-box}= [
    rectangle,
    rounded corners,
    text opacity=1,
    minimum height=1.75em,
    minimum width=5em,
    inner sep=2pt,
    align=center,
    fill opacity=.5,
]

\tikzstyle{leaf}=[
    my-box,
    minimum height=1.75em,
    fill opacity=0,
    text=black,
    align=left,
    text ragged,
    font=\small,
    inner xsep=2.5pt,
    inner ysep=4pt,
]

\forestset{
    rootblock/.style={draw=gray!65, fill=gray!8},
    levelone/.style={for tree={draw=L1!80!black, fill=L1!6}, text=L1!85!black, font=\small\bfseries},
    leveltwo/.style={for tree={draw=L2!80!black, fill=L2!6}, text=L2!80!black, font=\small\bfseries},
    levelthree/.style={for tree={draw=L3!80!black, fill=L3!8}, text=L3!85!black, font=\small\bfseries},
    levelfour/.style={for tree={draw=L4!80!black, fill=L4!8}, text=L4!85!black, font=\small\bfseries},
    levelfive/.style={for tree={draw=L5!80!black, fill=L5!7}, text=L5!85!black, font=\small\bfseries},
    levelsix/.style={for tree={draw=L6!80!black, fill=L6!6}, text=L6!80!black, font=\small\bfseries},
}

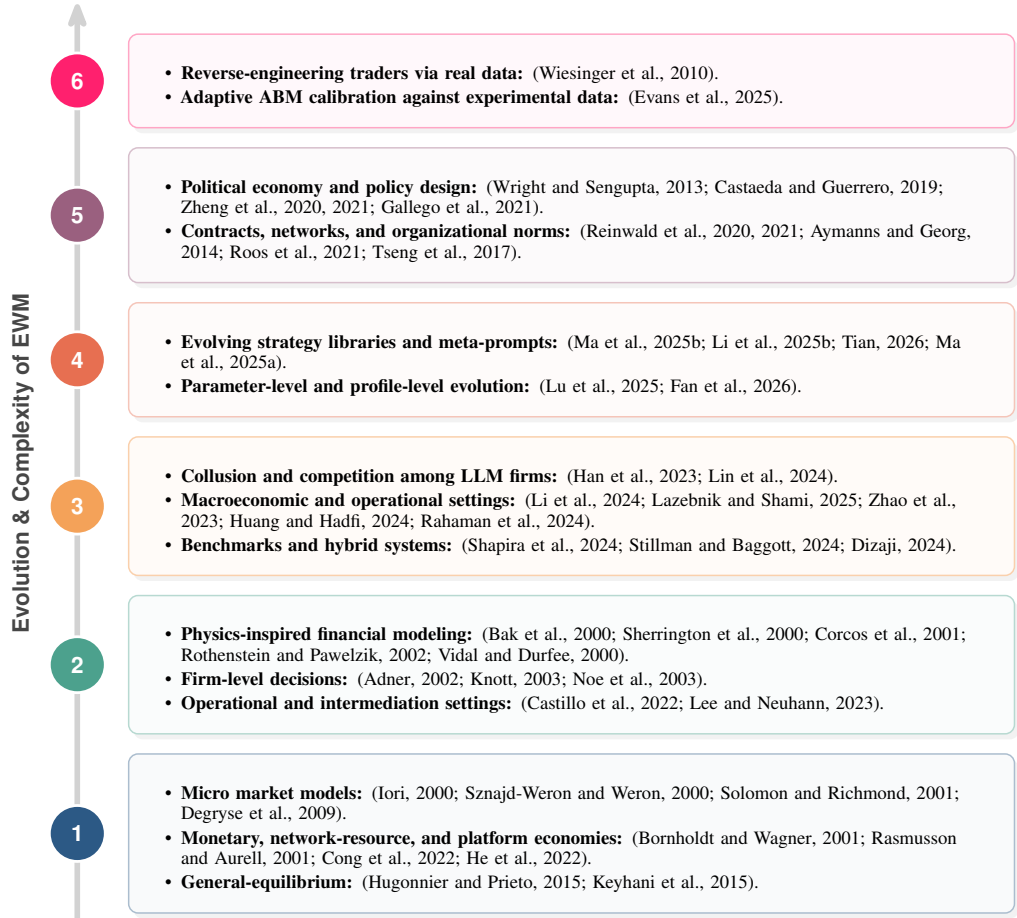
\begin{figure*}[th]
\centering
\resizebox{0.96\textwidth}{!}{%
\begin{tikzpicture}[
    node distance=0.35cm,
    spine/.style={line width=3pt, gray!35},
    dot/.style={circle, minimum size=1.05cm, inner sep=0pt, font=\bfseries\large\sffamily, text=white},
    card/.style={
        rectangle, rounded corners=5pt, thick,
        text width=15.5cm, inner xsep=16pt, inner ysep=11pt,
        drop shadow={opacity=0.10, shadow xshift=2.5pt, shadow yshift=-3pt}
    }
]

\definecolor{L1}{RGB}{44, 89, 133}
\definecolor{L2}{RGB}{76, 161, 141}
\definecolor{L3}{RGB}{244, 162, 89}
\definecolor{L4}{RGB}{231, 111, 81}
\definecolor{L5}{RGB}{154, 96, 127}
\definecolor{L6}{RGB}{255, 32, 110}

\node[card, draw=L1!40, fill=L1!3] (L1) {
    \begin{itemize}[leftmargin=1.2em, itemsep=2pt, parsep=0pt, topsep=2pt]
        \item \textbf{Micro market models:} \citep{iori2000scaling, sznajdweron2000simple, solomon2001power, degryse2009dynamic}.
        \item \textbf{Monetary, network-resource, and platform economies:} \citep{bornholdt2001stability, rasmusson2001price, cong2022token, he2022treasury}.
        \item \textbf{General-equilibrium:} \citep{hugonnier2015asset, keyhani2015theory}.
    \end{itemize}
};
\node[card, draw=L2!40, fill=L2!3, above=of L1] (L2) {
    \begin{itemize}[leftmargin=1.2em, itemsep=2pt, parsep=0pt, topsep=2pt]
        \item \textbf{Physics-inspired financial modeling:} \citep{bak2000money, sherrington2000statistical, corcos2001imitation, rothenstein2002evolution, vidal2000predicting}.
        \item \textbf{Firm-level decisions:} \citep{adner2002when, knott2003persistent, noe2003corporate}.
        \item \textbf{Operational and intermediation settings:} \citep{castillo2022designing, lee2023collateral}.
    \end{itemize}
};
\node[card, draw=L3!50, fill=L3!4, above=of L2] (L3) {
    \begin{itemize}[leftmargin=1.2em, itemsep=2pt, parsep=0pt, topsep=2pt]
        \item \textbf{Collusion and competition among LLM firms:} \citep{han2023guinea, lin2024strategic}.
        \item \textbf{Macroeconomic and operational settings:} \citep{li2024econagent, lazebnik2025investigating, zhao2023competeai, huang2024how, rahaman2024language}.
        \item \textbf{Benchmarks and hybrid systems:} \citep{shapira2024glee, stillman2024neuro, dizaji2024incentives}.
    \end{itemize}
};
\node[card, draw=L4!40, fill=L4!3, above=of L3] (L4) {
    \begin{itemize}[leftmargin=1.2em, itemsep=2pt, parsep=0pt, topsep=2pt]
        \item \textbf{Evolving strategy libraries and meta-prompts:} \citep{ma2025agent, li2025quantagents, tian2026prompt, ma2025think}.
        \item \textbf{Parameter-level and profile-level evolution:} \citep{lu2025aligning, fan2026aivilization}.
    \end{itemize}
};
\node[card, draw=L5!40, fill=L5!3, above=of L4] (L5) {
    \begin{itemize}[leftmargin=1.2em, itemsep=2pt, parsep=0pt, topsep=2pt]
        \item \textbf{Political economy and policy design:} \citep{wright2013modeling, castaeda2019importance, zheng2020ai, zheng2021ai, gallego2021data}.
        \item \textbf{Contracts, networks, and organizational norms:} \citep{reinwald2020agent, reinwald2021limited, aymanns2014contagious, roos2021effects, tseng2017humans}.
    \end{itemize}
};
\node[card, draw=L6!40, fill=L6!3, above=of L5] (L6) {
    \begin{itemize}[leftmargin=1.2em, itemsep=2pt, parsep=0pt, topsep=2pt]
        \item \textbf{Reverse-engineering traders via real data:} \citep{wiesinger2010reverse}.
        \item \textbf{Adaptive ABM calibration against experimental data:} \citep{evans2025adage}.
    \end{itemize}
};

\coordinate (spineBot) at ([xshift=-0.95cm, yshift=-0.15cm]L1.south west);
\coordinate (spineTop) at ([xshift=-0.95cm, yshift=0.55cm]L6.north west);
\draw[spine, -{Stealth[scale=1.1, round]}] (spineBot) -- (spineTop);

\foreach \n/\c/\num in {L1/L1/1, L2/L2/2, L3/L3/3, L4/L4/4, L5/L5/5, L6/L6/6} {
    \node[dot, fill=\c, draw=white, line width=1.5pt]
        at (\n.west -| spineBot) {\num};
}

\node[rotate=90, anchor=south, font=\bfseries\large\sffamily, text=gray!55!black]
    at ([xshift=-0.7cm]$(spineBot)!0.5!(spineTop)$) {Evolution \& Complexity of EWM};

\end{tikzpicture}
}
\caption{Representative systems by EWM level.}
\label{fig:representative_systems_by_level}
\end{figure*}

\noindent\textbf{Paper retrieval.}
Candidate papers are collected from two complementary sources between 1950 and April 2026: arXiv and journals in the UTD-24 list indexed by Web of Science. For arXiv, we search eight categories covering economics, finance, AI, multi-agent systems, machine learning, statistics, and quantitative finance. For Web of Science, we restrict the search to the UTD-24 journal list to obtain a peer-reviewed comparison set from leading business and economics journals. We apply an intersection-based title-and-abstract keyword filter, requiring each paper to contain at least one economic or financial term and at least one modeling, simulation, agent, reinforcement-learning, or world-model term. After merging and deduplication, the stage-one candidate pool contains 7,836 papers, including 6,008 unique arXiv papers and 1,828 UTD-24 papers. The complete keyword list and search protocol are reported in Appendix~\ref{app:Data}.

\noindent\textbf{Paper classification.}
We then classify the candidate papers through a two-stage LLM-assisted screening process. The first stage uses \texttt{GPT-5.4-mini} for title-and-abstract screening. This stage retains papers that show plausible EWM structure, including economic agents, endogenous outcomes from interaction, and dynamic feedback or repeated transitions over time. It reduces the pool to 794 arXiv papers and 87 UTD-24 papers for full-text review. The second stage uses \texttt{GPT-5.5} for full-text classification. Each paper is first validated against the EWM definition. Validated EWM papers are then assigned to the highest supported capability level from L1 to L6 based on concrete implementation evidence. Borderline and low-confidence cases are flagged for manual author verification. This process yields a final corpus of 737 validated EWM papers. More details are provided in Appendix~\ref{app:cls}.

\noindent\textbf{Representative literature.}
Figure~\ref{fig:representative_systems_by_level} lists representative papers for each of the six EWM levels. L1 includes fixed rule-based agent worlds, spanning micro market models; monetary, network-resource, and platform economies; and general-equilibrium or asset-pricing models. L2 covers adaptive rule-based agent worlds, encompassing physics-inspired financial modeling, firm-level decisions, and operational and intermediation settings. L3 captures LLM-based autonomous agent worlds, including collusion and competition among LLM firms, macroeconomic and operational settings, and benchmarks and hybrid systems. L4 includes self-evolving agent worlds, where agents accumulate evolving strategy libraries and meta-prompts, together with parameter-level and profile-level evolution. L5 moves from agent-side adaptation to world-side evolution, covering political economy and policy design as well as contracts, networks, and organizational norms. L6 contains sim-to-real economic twins, comprising early prototypes that reverse-engineer traders from real data and adaptively calibrate agent-based models against experimental data. A fuller discussion of the representative papers is provided in Appendix~\ref{app:representative_papers}.

\subsection{Current Status of EWM Research}
\label{subsec:corpus-screening}
\begin{figure}[t]
    \centering
    \includegraphics[width=1\columnwidth]{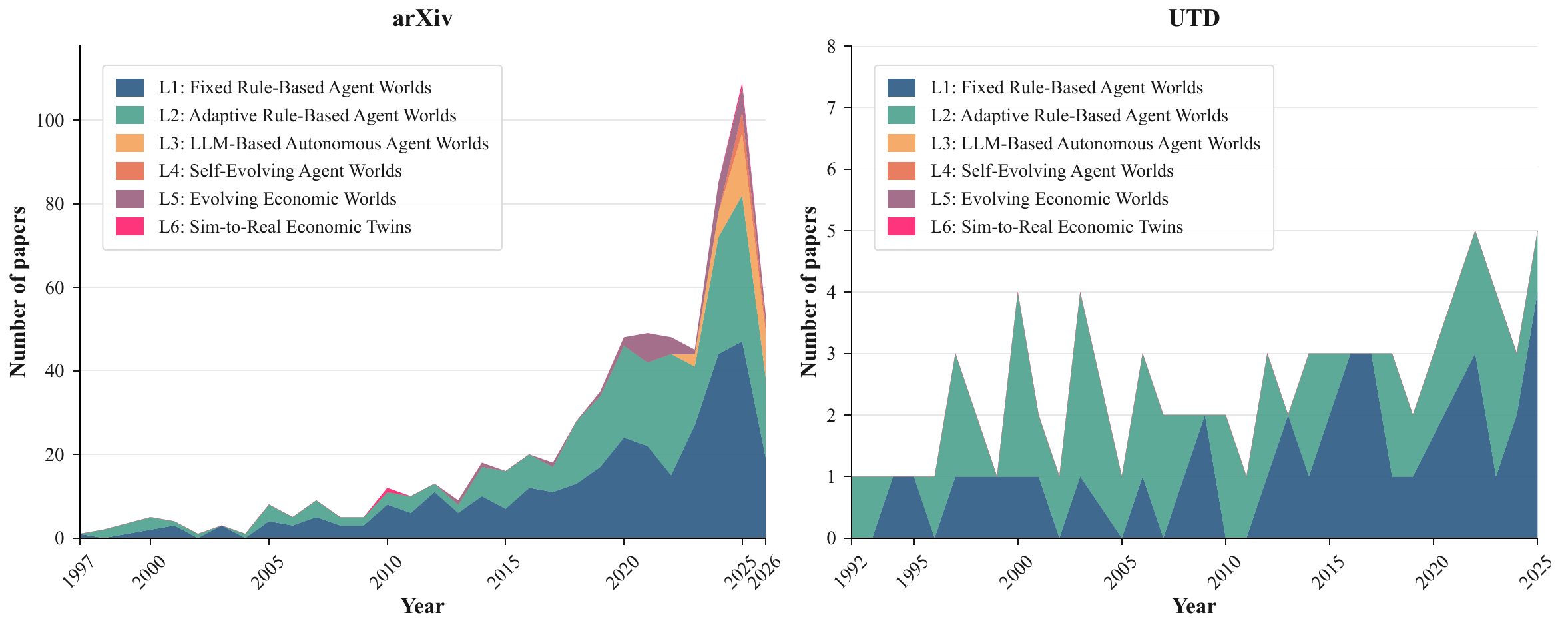}
    \caption{Literature landscape of EWM papers across years, source types, and capability levels.}
    \label{fig:six_level_bubble_timeline}
\end{figure}

Starting from \textbf{7,836} candidate papers, the pipeline yields \textbf{737} validated EWM papers. Figure~\ref{fig:six_level_bubble_timeline} summarizes the resulting distribution across years, source types, and capability levels. The classification reveals a few patterns.

\begin{itemize}[leftmargin=*]
    \item \textbf{Agent intelligence has become the main frontier of recent progress.}
    The arXiv panel shows a clear acceleration of EWM-related research in recent years. Before the early 2010s, the annual number of papers is small and dominated by L1--L2 systems. After 2018, the volume grows steadily, and around 2024--2025 it rises sharply. Most of this growth is concentrated in L1--L3, especially systems with adaptive rule-based agents and LLM-based autonomous agents. This pattern suggests that recent progress is driven mainly by richer agent intelligence, including learning, reasoning, memory, and language-based decision-making. However, the expansion of agent capability has not yet been matched by a comparable expansion of evolving economic worlds. In the figure, L4--L6 remain visually small relative to the large L1--L3 areas.
    \item \textbf{Business and economics journals remain concentrated at lower levels.}
    The UTD-24 panel shows a much smaller and more stable body of work. Annual counts are low throughout the sample period, and the distribution is concentrated almost entirely in L1 and L2. This reflects the longer tradition of economics, finance, management, and operations research in building rule-based or adaptive agent-based models with explicit mechanisms and economic interpretation. Compared with arXiv, UTD-24 papers contribute stronger economic discipline and clearer institutional structure, but they rarely incorporate LLM-based autonomous agents, persistent agent self-evolution, or sim-to-real correction. The contrast between the two panels therefore reveals a division of labor: arXiv pushes new AI capabilities, while business and economics journals provide more economically grounded modeling traditions.
    \item \textbf{Higher-level EWMs remain rare.}
    The figure also shows that L4--L6 systems occupy only thin areas in the overall literature landscape. L4 systems with self-evolving agents appear only recently and remain much less common than L1--L3 systems. L5 systems with endogenous institutional or rule evolution are even rarer, indicating that most simulations still treat the economic environment as a fixed stage rather than an evolving object. L6 sim-to-real economic twins are nearly absent, with only very limited evidence of repeated correction against real-world observations. This distribution highlights the main frontier of EWM research. The field has made visible progress in building more capable agents, but it has not yet solved the harder systems problem of integrating self-evolving agents, evolving institutions, and online real-world alignment within the same economic world.
\end{itemize}

\section{Implementation Architecture for EWM Systems}\label{sec:implementation}
This section first identifies the core modules of an EWM and then turns each module into a protocol interface. Section~\ref{subsec:core-components} gives the component map. Section~\ref{sec:reusable_implementation} specifies the reusable interfaces for agents, mechanisms, co-evolution, and real-time alignment. 

\subsection{Core Components of EWMs}
\label{subsec:core-components}

\begin{figure}[ht]
    \centering
    \includegraphics[width=0.9\columnwidth]{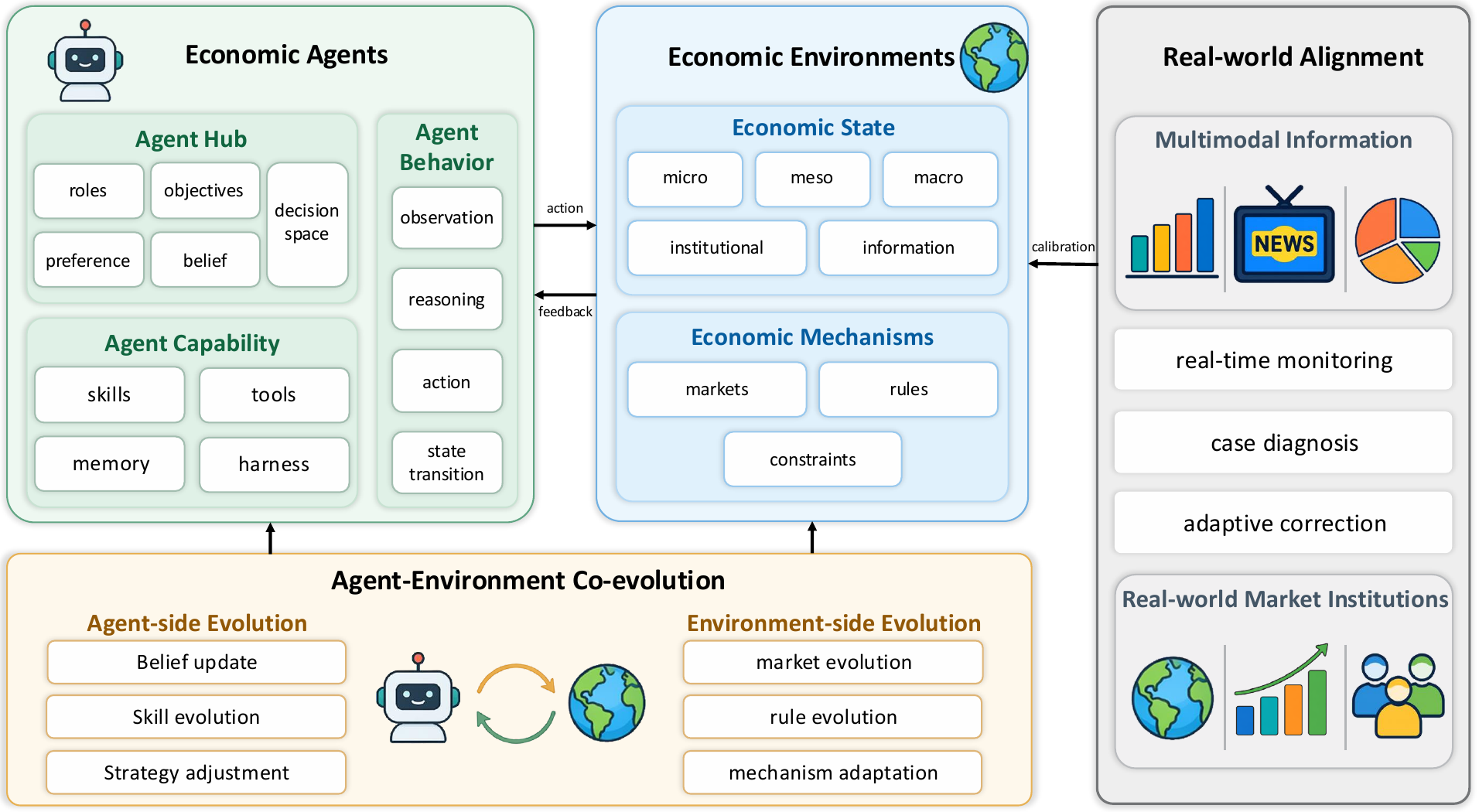}
    \caption{Core components of an Economic World Model.}
    \label{fig:components}
\end{figure}

Figure~\ref{fig:components} summarizes the core components of an Economic World Model. The figure is best read as a module map rather than a simulator blueprint: an EWM implementation can choose different domains, agent backends, mechanisms, datasets, and evaluation targets, but it should preserve the same separation between economic actors, the executable world, endogenous co-evolution, and empirical alignment.

\begin{itemize}[leftmargin=*]
\item \noindent\textbf{Agents.}
Agents are executable economic actors. Each agent has a role, objective, internal state, information channels, action space, constraints, beliefs, memory, tools, and skills. In the protocol, an agent is not merely a text generator or a policy function. It is a stateful decision process that observes permitted signals, forms expectations, proposes typed economic actions, and receives feedback from the world.

\item \noindent\textbf{Economic  environments.}
The economic world is the environment-style runtime that stores state and executes transitions. It publishes heterogeneous observations, receives actions, applies constraints, schedules actions when order matters, clears mechanisms, updates accounts and contracts, emits rewards and diagnostics, and records event logs. This world layer is where prices, allocations, inventories, contracts, exposures, and institutional states are generated endogenously.

Mechanisms are the institutional rules that mediate feasible actions: auctions, order books, matching rules, bargaining protocols, credit-allocation rules, tax schedules, settlement systems, and policy rules. Constraints define the feasibility boundary before these mechanisms execute, including budgets, inventories, role permissions, accounting identities, regulatory limits, and safety bounds.

\item \noindent\textbf{Co-evolution.}
Co-evolution is the bidirectional improvement loop between agents and the world. Better agents improve the world by producing harder trajectories, exposing missing rules, stress-testing mechanisms, and revealing unrealistic dynamics. Better worlds improve agents by providing structured feedback, counterfactual rollouts, rewards, constraint signals, memories, and stress cases.

\item \noindent\textbf{Real-time alignment.}
Alignment anchors the artificial economy to empirical, institutional, safety, or task-specific evidence during rollout. It corrects infeasible or dangerous actions, monitors drift and instability, attributes discrepancies to agents, states, mechanisms, or rules, and records interventions and versioned corrections.
\end{itemize}

\subsection{A Reusable Implementation Protocol for EWMs}\label{sec:reusable_implementation}

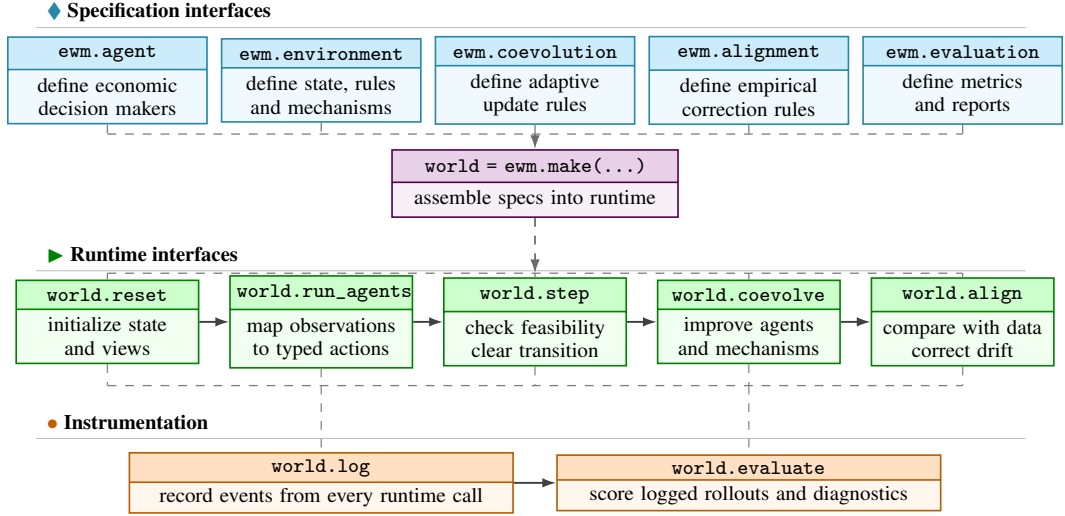
\begin{figure}[ht]
\centering
\resizebox{\linewidth}{!}{%
\begin{tikzpicture}[
    font=\scriptsize,
    >=Latex,
    spec/.style={
        rectangle split,
        rectangle split parts=2,
        rectangle split part fill={cyan!18,cyan!5},
        draw=cyan!60!black,
        line width=0.55pt,
        align=center,
        text width=2.20cm,
        minimum height=0.98cm,
        inner xsep=2.5pt,
        inner ysep=2.5pt
    },
    worldbox/.style={
        rectangle split,
        rectangle split parts=2,
        rectangle split part fill={violet!18,violet!6},
        draw=violet!65!black,
        line width=0.55pt,
        align=center,
        text width=3.20cm,
        minimum height=0.94cm,
        inner xsep=3pt,
        inner ysep=2.5pt
    },
    runtime/.style={
        rectangle split,
        rectangle split parts=2,
        rectangle split part fill={green!20,green!6},
        draw=green!50!black,
        line width=0.55pt,
        align=center,
        text width=2.00cm,
        minimum height=1.02cm,
        inner xsep=2.5pt,
        inner ysep=2.5pt
    },
    instrument/.style={
        rectangle split,
        rectangle split parts=2,
        rectangle split part fill={orange!24,orange!7},
        draw=orange!72!black,
        line width=0.55pt,
        align=center,
        text width=4.40cm,
        minimum height=0.72cm,
        inner xsep=2.5pt,
        inner ysep=1.8pt
    },
    header/.style={
        font=\scriptsize\bfseries,
        align=left,
        text=black
    },
    bus/.style={draw=black!50, line width=0.45pt, dashed},
    arrow/.style={-{Latex[length=1.65mm,width=1.15mm]}, line width=0.65pt, draw=black!75},
    dasharrow/.style={-{Latex[length=1.65mm,width=1.15mm]}, line width=0.6pt, dashed, draw=black!58},
    trace/.style={draw=black!45, line width=0.45pt, dashed}
]

\node[header, anchor=west] at (-0.85,4.12) {{\color{cyan!60!black}\(\blacklozenge\)} Specification interfaces};
\draw[black!25, line width=0.35pt] (-0.85,3.93) -- (10.95,3.93);

\node[spec] (agent) at (0,3.30) {
    \texttt{ewm.agent}
    \nodepart{two}
    define economic\\
    decision makers
};

\node[spec] (environment) at (2.55,3.30) {
    \texttt{ewm.environment}
    \nodepart{two}
    define state, rules\\
    and mechanisms
};

\node[spec] (coevolution) at (5.10,3.30) {
    \texttt{ewm.coevolution}
    \nodepart{two}
    define adaptive\\
    update rules
};

\node[spec] (alignment) at (7.65,3.30) {
    \texttt{ewm.alignment}
    \nodepart{two}
    define empirical\\
    correction rules
};

\node[spec] (evaluation) at (10.20,3.30) {
    \texttt{ewm.evaluation}
    \nodepart{two}
    define metrics\\
    and reports
};

\node[worldbox] (make) at (5.10,2.07) {
    \texttt{world = ewm.make(...)}
    \nodepart{two}
    assemble specs into runtime
};

\coordinate (specbusL) at (0,2.66);
\coordinate (specbusR) at (10.20,2.66);
\draw[bus] (specbusL) -- (specbusR);
\foreach \n in {agent,environment,coevolution,alignment,evaluation} {
    \draw[bus] (\n.south) -- (\n.south |- specbusL);
}
\draw[dasharrow] (5.10,2.66) -- (make.north);

\node[header, anchor=west] at (-0.85,1.22) {{\color{green!50!black}\(\blacktriangleright\)} Runtime interfaces};
\draw[black!25, line width=0.35pt] (-0.85,1.03) -- (10.95,1.03);

\node[runtime] (reset) at (0,0.42) {
    \texttt{world.reset}
    \nodepart{two}
    initialize state\\
    and views
};

\node[runtime] (runagents) at (2.55,0.42) {
    \texttt{world.run\_agents}
    \nodepart{two}
    map observations\\
    to typed actions
};

\node[runtime] (step) at (5.10,0.42) {
    \texttt{world.step}
    \nodepart{two}
    check feasibility\\
    clear transition
};

\node[runtime] (coevolve) at (7.65,0.42) {
    \texttt{world.coevolve}
    \nodepart{two}
    improve agents\\
    and mechanisms
};

\node[runtime] (align) at (10.20,0.42) {
    \texttt{world.align}
    \nodepart{two}
    compare with data\\
    correct drift
};

\coordinate (runtimebusL) at (0,1.00);
\coordinate (runtimebusR) at (10.20,1.00);
\draw[bus] (runtimebusL) -- (runtimebusR);
\foreach \n in {reset,runagents,step,coevolve,align} {
    \draw[bus] (\n.north) -- (\n.north |- runtimebusL);
}
\draw[dasharrow] (make.south) -- (5.10,1.00);

\draw[arrow] (reset) -- (runagents);
\draw[arrow] (runagents) -- (step);
\draw[arrow] (step) -- (coevolve);
\draw[arrow] (coevolve) -- (align);

\node[header, anchor=west] at (-0.85,-0.78) {{\color{orange!75!black}\(\bullet\)} Instrumentation};
\draw[black!25, line width=0.35pt] (-0.85,-0.97) -- (10.95,-0.97);

\coordinate (traceL) at (0,-0.34);
\coordinate (traceR) at (10.20,-0.34);
\draw[bus] (traceL) -- (traceR);
\foreach \n in {reset,runagents,step,coevolve,align} {
    \draw[trace] (\n.south) -- (\n.south |- traceL);
}

\node[instrument] (logruntime) at (2.55,-1.50) {
    \texttt{world.log}
    \nodepart{two}
    record events from every runtime call
};

\node[instrument] (evalruntime) at (7.65,-1.50) {
    \texttt{world.evaluate}
    \nodepart{two}
    score logged rollouts and diagnostics
};

\draw[trace] (2.55,-0.34) -- (logruntime.north);
\draw[trace] (7.65,-0.34) -- (evalruntime.north);
\draw[arrow] (logruntime) -- (evalruntime);

\end{tikzpicture}%
}
\caption{Reusable implementation protocol for EWMs.}
\label{fig:reusable-ewm-protocol}
\end{figure}

The previous subsection identifies the core modules of an EWM. We now turn these modules into a reusable implementation protocol. The protocol separates two kinds of interfaces. The first is the \emph{specification interface}, which declares the configurable components of the world: agents, environments, co-evolution rules, real-world alignment targets, and evaluation metrics. The second is the \emph{runtime interface}, which defines how these components interact during rollout: agents generate typed actions, the environment advances the state, co-evolution updates adaptive components, online alignment anchors the world to external evidence, and evaluation reads the generated trajectory. 

Figure~\ref{fig:reusable-ewm-protocol} summarizes the reusable EWM protocol. 
The public runtime boundary is intentionally compact. 
To begin, users construct the world with \textbf{\texttt{ewm.make}} and initialize it with \textbf{\texttt{world.reset}}. 
During execution, they generate agent actions with \textbf{\texttt{world.run\_agents}} and advance the economy with \textbf{\texttt{world.step}}. 
Finally, users update adaptive components with \textbf{\texttt{world.coevolve}}, align the simulated world with real data through \textbf{\texttt{world.align}}, and evaluate trajectories with \textbf{\texttt{world.evaluate}}. 
Behind these stable calls lies a modular execution layer: 
agent interfaces govern perception and action generation; 
environment interfaces handle constraints and economic transitions; 
co-evolution and alignment interfaces enable adaptive and empirically grounded evolution; 
and evaluation interfaces provide trajectory-level diagnostics. The following subsections instantiate each part of this protocol.

\subsubsection{Economic Agents}
\label{subsec:economic_agents}
\noindent\textbf{Agent specification interface.}
Each agent is configured to represent a specific economic role and to expose a role-specific decision interface. Figure~\ref{fig:agent-api-example} gives a minimal configuration for a household agent in a foreign-exchange setting. The specification declares the agent's cash and foreign-exchange (FX) holdings, beliefs about the exchange rate, public and private information channels, and admissible actions, including buying FX, selling FX, or holding.

\begin{figure}[ht]
\centering
\begin{minipage}{0.9\textwidth}
\begin{lstlisting}[style=ewmpython]
import economic_world_model as ewm

agent = ewm.agent(
        role="household",
        objective="Hold cash and FX while respecting budget constraints.",
        state_variables=["cash", "fx_inventory", "belief"],
        information_channels={
            "public": ["exchange_rate", "transaction_volume"],
            "private": ["cash", "fx_inventory"]
        },
        action_space=["buy_fx", "sell_fx", "hold"],
        tools=["budget_calculator", "fx_quote_lookup"],
        constraints=["budget", "inventory", "role_permission"],
        memory_window=4
        )
\end{lstlisting}
\end{minipage}
\caption{Example specification of an agent in an EWM.}
\label{fig:agent-api-example}
\end{figure}

The interface organizes an economic agent into three groups of attributes.
\begin{itemize} 
\item \textbf{Attributes.} \texttt{role} and \texttt{objective} specify the agent's economic identity and decision goal. \texttt{state\_variables}, \texttt{belief}, and \texttt{information\_channels} describe its internal economic state, subjective expectations, and access to public and private information. 
\item \textbf{Capabilities.} \texttt{action\_space} defines the economic actions available to the agent and each action may include the direction, price, and volume related to the transaction. \texttt{Tools} provide auxiliary capabilities for reasoning, calculation, information retrieval, and interaction with the economic environment. 
\item \textbf{Boundaries.} \texttt{constraints} ensure that generated actions satisfy budget, inventory, institutional, and role-specific requirements before entering the world transition. \texttt{memory\_window} limits the amount of historical context available to the agent, representing a boundary on its information processing and decision-making. \end{itemize}

\noindent\textbf{Runtime action-generation interface.}
Once agents are specified, they are invoked through a runtime action-generation interface. As shown in Figure~\ref{fig:agent-action-generation}, the public call \texttt{world.run\_agents(state)} applies the agent-side decision loop to all agents and maps the current economic state to a profile of typed economic actions. Inside this loop, each agent first constructs a role-specific observation from its information channels. Next, it selects an action type from its predefined \texttt{action\_space} and fills in the required action parameters. Finally, it validates the action against its constraints.

\begin{figure}[ht!]
\centering
\begin{minipage}{0.9\textwidth}
\begin{lstlisting}[style=ewmpython]
world = ewm.make(agents=[agent_1,agent_2,...] )

# Public protocol call: state_t -> action profile A_t.
actions = world.run_agents(state, parallel=True)

# Internal decision loop for one agent. 
def generate_action(agent, state):           
    observation = agent.observe(state, channels=agent.information_channels) 
    action_type = agent.select_action(observation, agent.objective, agent.belief, 
                                      agent.memory, agent.action_space) 
    action = agent.instantiate_action(action_type, observation, agent.tools) 
    action = agent.validate_action(action, constraints=agent.constraints)

    return action
\end{lstlisting}
\end{minipage}
\caption{Agent action-generation protocol.}
\label{fig:agent-action-generation}
\end{figure}

\subsubsection{Economic Environment}
\label{subsec:economic_environment}

The environment protocol defines the executable world around agents. It stores the current economic state, publishes observations, checks feasibility, optionally orders actions, and executes mechanisms. A mechanism is the institutional rule that turns feasible actions and the current economic state into the next economic state. 

\begin{figure}[ht]
\centering
\begin{minipage}{0.9\textwidth}
\begin{lstlisting}[style=ewmpython]
environment = ewm.environment(
        state=ewm.state(
            variables={
                "exchange_rate": 1.00,
                "volume": 1000.0,
                "volatility": 0.02,
            },
            accounts={
                "household": {"cash": 1000.0, "fx_inventory": 0.0},
                "firm": {"cash": 3000.0, "fx_need": 500.0},
                "bank": {"cash": 10000.0, "fx_inventory": 2000.0},
            }
        ),
        constraints=ewm.constraints(
            rules=["budget", "inventory", "role_permission", "exposure_limit"],
            violation_policy="reject_and_log"
        ),
        scheduler=ewm.scheduler(),
        mechanism=ewm.mechanism(
            type="batch_clearing",
            participants=["household", "firm", "bank"],
            input_actions=["buy_fx", "sell_fx", "quote_bid", "quote_ask"],
            pricing_rule="uniform_clearing",
            settlement_rule="cash_asset_delivery"
        )
    )

world = ewm.make(name="FXMarket-v0",
                 agents=agents,
                 environment=environment
)
\end{lstlisting}
\end{minipage}
\caption{Minimal environment interface for EWMs. Continuing Figure~\ref{fig:agent-api-example}, this code adds the state, constraints, scheduler, and clearing mechanism needed to instantiate a runnable world.}
\label{fig:mechanism-api-example}
\end{figure}

\noindent\textbf{Definition and initialization.}
Environment is configured to represent an executable economic setting and to expose a mechanism-specific transition interface. Continuing the foreign-exchange example, Figure~\ref{fig:mechanism-api-example} defines the market state, role-level accounts, feasibility constraints, action scheduler, and clearing mechanism. The environment does not decide what agents want to do. Instead, it specifies what the world contains and how feasible actions are transformed into prices, allocations, settlements, and the next economic state.

The interface organizes an economic environment into three groups of attributes. \begin{itemize} 
\item \textbf{State.} \texttt{state} defines the current economic world, including aggregate market variables such as exchange rates, market depth, and volatility, as well as role-level accounts such as cash, foreign-exchange inventory, and demand. 
\item \textbf{Feasibility.} \texttt{constraints} specify which submitted actions are admissible under budget, inventory, role-permission, and exposure rules.
\item \textbf{Ordering.} \texttt{scheduler} defines how actions are ordered when the timing or priority of actions matters. 
\item \textbf{Mechanism and settlement.} \texttt{mechanism} defines the institutional rule that mediates feasible actions. It specifies who participates, which action types can enter the mechanism, how prices or allocations are computed, and how cash and assets are settled. 
\end{itemize}

\noindent\textbf{Runtime state-transition interface.}
Once agents have produced an action profile, the environment is invoked through the runtime state-transition interface. Figure~\ref{fig:mechanism-execution-example} illustrates this boundary by showing both the public call and the internal transition loop. The public call \texttt{world.step(actions)} maps the current economic state and the submitted actions to the next economic state. Inside this loop, the environment first checks whether actions satisfy feasibility constraints. It then orders feasible actions when the mechanism requires timing or priority rules. The mechanism clears the market by computing prices, allocations, and settlements. Finally, the environment updates state variables, account balances, and transition logs. This interface does not ask agents to make new decisions; it only executes the economic consequences of the actions already submitted.

\begin{figure}[ht]
\centering
\begin{minipage}{0.9\textwidth}
\begin{lstlisting}[style=ewmpython]
# Public protocol call: (state_t, action profile A_t) -> state_{t+1}.
next_state = world.step(actions)

# Internal transition loop for the environment.
def transition_state(environment, state, actions):
    # Check whether submitted actions satisfy feasibility rules.
    feasible_actions = environment.check_constraints(actions, state=state)
    
    # Order feasible actions when timing or priority matters.
    ordered_actions = environment.schedule_actions(feasible_actions, state=state)
    
    # Execute the mechanism to compute prices, allocations, and settlement.
    next_state = environment.mechanism.clear(state=state, actions=ordered_actions)
    
    # Record actions, outcomes, and the resulting state.
    environment.update_log()

    return next_state
\end{lstlisting}
\end{minipage}
\caption{Environment state-transition protocol. The environment checks submitted actions, schedules feasible actions, executes the market mechanism, and updates the economic state.}
\label{fig:mechanism-execution-example}
\end{figure}

\subsubsection{Agent--Environment Co-evolution}
\label{subsec:agent_environment_coevolution}

\noindent\textbf{Co-evolution specification interface.}
Agent--environment co-evolution defines how agents and environments adapt through repeated economic interactions. In an EWM, agents do not simply act once under fixed preferences, and environments do not remain passive containers with frozen parameters. Instead, agents update their beliefs, memory, or policies from feedback generated by the environment, such as realized prices, allocations, rewards, constraint feedback, and the next economic state. Conversely, the environment recalibrates market parameters, volatility, or mechanism parameters from agent-generated behavioral signals. As shown in Figure~\ref{fig:coevolution-api-example}, this interface specifies which components on each side are allowed to evolve and which feedback signals drive their updates.

\begin{figure}[ht]
\centering
\begin{minipage}{0.9\textwidth}
\begin{lstlisting}[style=ewmpython]
coevolution = ewm.coevolution(
        agent_updates=ewm.agent_updates(
            targets=["belief", "memory", "policy"],
            signals=["observation", "realized_outcome", "reward"],
        ),
        environment_updates=ewm.environment_updates(
            targets=["mechanism_parameters"],
            signals=["trading_volume", "price_error", "constraint_violations"],
        )
    )

world = ewm.make("FXMarket-v0",
                 agents=agents,
                 environment=environment,
                 coevolution=coevolution
)
\end{lstlisting}
\end{minipage}
\caption{Minimal co-evolution interface for EWMs. The specification defines which agent-side and environment-side components can be updated, and which signals trigger updates.}
\label{fig:coevolution-api-example}
\end{figure}

\noindent\textbf{Runtime co-evolution interface.} After agents act and the environment produces the next state, the co-evolution interface updates the adaptive components of the world. Figure~\ref{fig:coevolution-execution-example} shows this runtime loop. Each period first calls \texttt{world.run\_agents(state)} to generate an action profile, and then calls \texttt{world.step(actions)} to execute the economic transition. The resulting \texttt{next\_state} contains the updated economic state as well as realized feedback from the environment, such as prices, allocations, account changes, and constraint feedback. The call \texttt{world.coevolve(state, next\_state)} then applies a two-way update: agents evolve from environment feedback, while the environment evolves from aggregate agent behavior. This interface does not replace agent decision-making or environment transition. Instead, it governs how both sides adapt through repeated interaction.

\begin{figure}[ht]
\centering
\begin{minipage}{0.9\textwidth}
\begin{lstlisting}[style=ewmpython]
state = world.reset(seed=42)

for t in range(T):
    actions = world.run_agents(state, parallel=True)
    next_state = world.step(actions)

    # Co-evolution update based on the realized transition.
    world.coevolve(state, actions, next_state)

    state = next_state

# Internal co-evolution loop.
def coevolve(world, state, next_state):
    # Agents evolve from feedback produced by the environment.
    agent_feedback = world.extract_environment_feedback(state, next_state)
    world.update_agents(agent_feedback)

    # The environment evolves from aggregate agent behavior.
    behavior_feedback = world.aggregate_agent_behavior(state, next_state)
    world.update_environment(behavior_feedback)

    return world
\end{lstlisting}
\end{minipage}
\caption{Agent--environment co-evolution protocol. Agents evolve from environment feedback, while the environment recalibrates itself from aggregate agent behavior.}
\label{fig:coevolution-execution-example}
\end{figure}

This co-evolution layer makes EWMs different from static simulations. In a static simulator, agents act under fixed rules and the environment only applies those rules. In an EWM, both sides can adapt under explicit controls. Agent adaptation captures learning, expectation revision, and strategic adjustment. Environment adaptation captures mechanism recalibration and rule refinement. This allows the EWM to represent economies as evolving systems rather than one-shot input--output mappings.

\subsubsection{Online Real-World Alignment}
\label{subsec:real_world_alignment}

\noindent\textbf{Real-world alignment specification interface.} Online real-world alignment defines how an EWM is repeatedly compared with external evidence and corrected when it drifts away from the real economy. Agent--environment co-evolution governs adaptation inside the simulated world, but such adaptation may become internally consistent while remaining externally misaligned. The alignment process therefore acts as a form of world-level supervision. Observed discrepancies provide learning signals for updating agent beliefs and behaviors, environment dynamics, and economic mechanisms. As these corrections are internalized, the EWM can progressively improve its open-loop simulation accuracy. As shown in Figure~\ref{fig:alignment-api-example}, the goal is not to overwrite the simulation with real observations, but to keep its evolving internal dynamics empirically anchored.

\begin{figure}[ht]
\centering
\begin{minipage}{0.9\textwidth}
\begin{lstlisting}[style=ewmpython]
alignment = ewm.alignment(
        data_sources=ewm.data_sources(
            streams=["exchange_rate", "trading_volume", "volatility"],
            frequency="daily"
        ),
        targets=["exchange_rate", "volume", "volatility"],
        metrics=["price_error", "volume_error", "volatility_error"],
        tolerance={
            "price_error": 0.02,
            "volume_error": 0.10,
            "volatility_error": 0.05
        },
        correction=ewm.correction(
            agent_targets=["belief", "state"],
            environment_targets=["mechanism_parameters"],
            policy="bounded_update"
        )
    )

world = ewm.make("FXMarket-v0",
                 agents=agents,
                 environment=environment,
                 coevolution=coevolution,
                 alignment=alignment
)
\end{lstlisting}
\end{minipage}
\caption{Minimal real-world alignment interface for EWMs. The specification defines external data streams, alignment targets, discrepancy metrics, tolerance thresholds, and bounded correction rules.}
\label{fig:alignment-api-example}
\end{figure}

The interface organizes real-world alignment into three groups of attributes.
\begin{itemize}
\item \textbf{Evidence and targets.} \texttt{data\_sources} defines the external signals used to anchor the world, such as exchange rates, trading volume, or volatility. \texttt{targets} specifies which simulated variables should be compared against these signals.
\item \textbf{Discrepancy measurement.} \texttt{metrics} and \texttt{tolerance} define how deviations between the simulated world and the real economy are measured, and when such deviations are large enough to require correction.
\item \textbf{Bounded correction.} \texttt{correction} specifies which agent-side or environment-side components may be adjusted. These updates are bounded so that alignment improves empirical fit without arbitrarily rewriting the economic mechanism.
\end{itemize}

\noindent\textbf{Runtime alignment interface.}
After the simulated world evolves, the alignment interface compares its generated state with real-world evidence. Figure~\ref{fig:alignment-execution-example} shows this runtime loop. Each period first generates actions, executes the environment transition, and applies co-evolution. The alignment module then retrieves external observations and measures the discrepancy between the simulated state and real data. If the discrepancy exceeds the specified tolerance, the system diagnoses the source of misalignment and applies bounded corrections to selected agent or environment components. This interface differs from co-evolution: co-evolution updates the world from its own realized interactions, while online alignment corrects the world using external empirical evidence.

\begin{figure}[ht]
\centering
\begin{minipage}{0.9\textwidth}
\begin{lstlisting}[style=ewmpython]
state = world.reset(seed=42)

for t in range(T):
    actions = world.run_agents(state, parallel=True)
    next_state = world.step(actions)
    world.coevolve(state, actions, next_state)

    # Real-world alignment based on external evidence.
    real_data = world.fetch_real_data(t)
    world.align(next_state, real_data)

    state = next_state

# Internal real-world alignment loop.
def align(world, simulated_state, real_data):
    discrepancy = world.measure_discrepancy(simulated_state, real_data)

    if discrepancy.exceeds_tolerance():
        diagnosis = world.diagnose_misalignment(discrepancy)
        corrections = world.plan_corrections(diagnosis)
        world.apply_corrections(corrections)

    return world
\end{lstlisting}
\end{minipage}
\caption{Online real-world alignment protocol. The world compares simulated states with external evidence, diagnoses discrepancies, and applies bounded corrections to selected agent or environment components.}
\label{fig:alignment-execution-example}
\end{figure}

This layer prevents EWMs from becoming closed, self-reinforcing simulations. Without real-world alignment, agents and environments may co-evolve in ways that are coherent inside the model but inconsistent with observed economic behavior. Online alignment provides the empirical correction channel required for building economic digital twins. It continuously compares simulated states with real-world signals, detects deviations, and applies bounded corrections to agent beliefs, market parameters, and mechanisms. In this sense, online alignment is the key step that turns an EWM from an offline simulator into an adaptive economic digital twin.

\subsubsection{Evaluation}
\label{subsec:evaluation}

\noindent\textbf{Layered evaluation stack.}
Evaluation measures an EWM as an executable economic world rather than as a single predictive model. Since an EWM contains agents, environments, co-evolution, and real-world alignment, its evaluation should be layered. A generated rollout contains agent decisions, environment transitions, adaptation updates, and alignment corrections. These traces allow us to evaluate not only whether the final simulated state is close to reality, but also where errors arise and which component causes them.

Table~\ref{tab:ewm-evaluation} summarizes the main evaluation targets. Agent-level metrics examine whether economic actors behave within their declared roles, objectives, beliefs, and constraints. Environment-level metrics verify whether the world correctly enforces feasibility rules, executes mechanisms, clears markets, and updates accounts. Co-evolution metrics evaluate whether repeated interaction improves adaptation without causing instability or drift. Real-world alignment metrics measure whether simulated states remain empirically anchored to observed economic evidence. Efficiency metrics evaluate whether the system remains scalable as the number of agents, markets, and rollout periods increases.

\begin{table}[ht]
\centering
\caption{Layered evaluation targets for Economic World Models.}
\label{tab:ewm-evaluation}
\scriptsize
\setlength{\tabcolsep}{4pt}
\renewcommand{\arraystretch}{1.18} 

\rowcolors{2}{white}{gray!8}

\begin{tabular}{m{0.20\textwidth} m{0.36\textwidth} m{0.32\textwidth}}
\toprule
\rowcolor{white}
\textbf{Layer} & \textbf{Evaluation focus} & \textbf{Example metrics} \\
\midrule

Agents 
& Whether agents act according to their roles, objectives, beliefs, and constraints. 
& Action validity, role consistency, belief calibration, behavioral diversity. \\

Environment 
& Whether the executable world correctly enforces constraints, mechanisms, and settlements. 
& Constraint violation rate, market-clearing error, accounting consistency, settlement correctness. \\

Co-evolution 
& Whether agents and environments adapt productively over repeated interaction. 
& Adaptation gain, stability, drift, policy change, mechanism recalibration quality. \\

Real-world alignment 
& Whether simulated states remain close to observed economic evidence. 
& State error, trend match, price error, volume error, volatility error, correction magnitude. \\

Efficiency 
& Whether the system remains scalable as agents, markets, and rollout length grow. 
& Runtime cost, memory usage, parallel efficiency, scalability with agent count. \\
\bottomrule
\end{tabular}
\end{table}

\noindent\textbf{Runtime evaluation protocol.}
At runtime, evaluation is applied to the trajectory generated by the world rather than to a single final state. Figure~\ref{fig:evaluation-execution-example} shows this evaluation boundary. The public call \texttt{world.evaluate(...)} consumes rollout logs, real-world evidence, and optional baselines, and returns a structured evaluation report. This call is read-only: it does not change agent beliefs, environment parameters, or alignment states. Instead, it diagnoses the generated trajectory across agent behavior, environment execution, co-evolution, real-world alignment, and efficiency.

\begin{figure}[ht]
\centering
\begin{minipage}{0.9\textwidth}
\begin{lstlisting}[style=ewmpython]
# Public evaluation call: trajectory -> evaluation report.
report = world.evaluate()

# Internal evaluation loop.
def evaluate(world, trajectory, real_data):
    agent_score = world.evaluator.score_agents(trajectory, real_data)
    env_score = world.evaluator.score_environment(trajectory, real_data)
    coev_score = world.evaluator.score_coevolution(trajectory, real_data)
    align_score = world.evaluator.score_alignment(trajectory, real_data)
    cost_score = world.evaluator.score_efficiency(trajectory, real_data)

    return (agent_score, env_score, coev_score, align_score, cost_score)
\end{lstlisting}
\end{minipage}
\caption{Runtime evaluation protocol. The evaluator reads the generated trajectory and produces component-level diagnostics.}
\label{fig:evaluation-execution-example}
\end{figure}

\section{The Engineering Path Toward Economic World Models}
\label{sec:roadmap}
The six-level hierarchy in Section~\ref{sec:toward} provides a capability-based view of what an EWM should be able to do. In this section, we take a complementary engineering view and trace the technical trajectory that leads to the notion of Economic World Models. We organize advances in AI applied to economics and finance into five engineering stages: feature engineering, data engineering, prompt engineering, context engineering, and environment engineering. This progression reveals that EWMs are not a theoretical leap from nowhere, but the logical outcome of a sustained engineering evolution.

Table~\ref{tab:waves_levels_matrix} summarizes how the five engineering waves support the six capability levels introduced above. The mapping is not one-to-one. Feature engineering and data engineering mainly determine how agents observe the economy. Feature engineering appears from L1, where rule-based agents act on designed economic indicators. Data engineering becomes important at L2, where adaptive agents update their strategies from signal-driven inputs. From L3 onward, prompt engineering and context engineering become central. Prompt engineering elicits role-conditioned reasoning from LLM-based agents. Context engineering gives these agents memory, tools, reflection, and eventually real-time information. These waves help move agents from autonomous reasoning at L3 to self-evolution at L4 and empirical alignment at L6. Environment engineering runs through all six levels. Its role deepens from specifying world rules at L1, to adding feedback at L2, supporting executable interaction at L3, enabling agent-environment coupling at L4 and L5, and grounding the system in real-world data at L6. This progression is what makes evolving economic worlds and sim-to-real twins possible.

\begin{table*}[h]
\centering
\caption{How engineering waves relate to EWM capability levels. Earlier waves mainly improve agent-side components of an EWM, while environment engineering is required to realize interactive and higher-level economic worlds.}
\label{tab:waves_levels_matrix}
\vspace{8pt}
\renewcommand{\arraystretch}{1.25}
\setlength{\tabcolsep}{5pt}
\newcolumntype{M}{>{\centering\arraybackslash}m{2.6cm}}
\resizebox{\textwidth}{!}{
\begin{tabular}{>{\centering\arraybackslash}m{2.8cm} M M M M M M >{\raggedright\arraybackslash}m{4.6cm}}
\toprule
\textbf{Engineering wave} 
& \textbf{\shortstack{L1\\Fixed rules}} 
& \textbf{\shortstack{L2\\Adaptive rules}} 
& \textbf{\shortstack{L3\\LLM agents}} 
& \textbf{\shortstack{L4\\Self-evolving}} 
& \textbf{\shortstack{L5\\Evolving worlds}} 
& \textbf{\shortstack{L6\\Real alignment}}
& \multicolumn{1}{>{\centering\arraybackslash}m{4.6cm}}{\textbf{Interpretation}} \\
\midrule
\rowcolor{teal!5}
\textbf{Feature}
& \cellcolor{teal!22}\textbf{observation}
& \cellcolor{teal!22}\textbf{observation}
& \cellcolor{teal!22}\textbf{observation}
& \cellcolor{teal!22}\textbf{observation}
& \cellcolor{teal!22}\textbf{observation}
& \cellcolor{teal!22}\textbf{observation}
& Design economic indicators as agents’ observation space. \\
\rowcolor{blue!5}
\textbf{Data}
& --
& \cellcolor{blue!18}\textbf{signals}
& \cellcolor{blue!32}\textbf{model knowledge}
& \cellcolor{blue!32}\textbf{model knowledge}
& \cellcolor{blue!32}\textbf{model knowledge}
& \cellcolor{blue!32}\textbf{model knowledge}
& Train stronger foundation models to enhance agent capabilities. \\
\rowcolor{olive!5}
\textbf{Prompt}
& --
& --
& \cellcolor{olive!28}\textbf{reasoning}
& \cellcolor{olive!28}\textbf{reasoning}
& \cellcolor{olive!28}\textbf{reasoning}
& \cellcolor{olive!28}\textbf{reasoning}
& Elicit role-conditioned reasoning and decisions. \\
\rowcolor{orange!5}
\textbf{Context}
& --
& --
& \cellcolor{orange!18}\textbf{memory/tool use}
& \cellcolor{orange!28}\textbf{reflection}
& \cellcolor{orange!28}\textbf{reflection}
& \cellcolor{orange!38}\textbf{real-time info}
& Equips agents with memory, retrieval, tools, reflective context, and real-time information. \\
\rowcolor{purple!5}
\textbf{Environment}
& \cellcolor{purple!12}\textbf{world rules}
& \cellcolor{purple!20}\textbf{feedback}
& \cellcolor{purple!28}\textbf{execution}
& \cellcolor{purple!36}\textbf{interaction}
& \cellcolor{purple!36}\textbf{interaction}
& \cellcolor{purple!44}\textbf{alignment}
& Builds economic worlds where agents act under rules, feedback, and reality alignment. \\
\bottomrule
\end{tabular}
}
\end{table*}

\subsection{Feature Engineering: Designing Observation Signals}
Feature engineering marks the first major wave of AI in economics. Its core workflow is to translate economic intuition and domain knowledge into informative signals, and then use machine learning (ML) models to predict economic outcomes such as asset returns, credit risk, and fraud. In empirical asset pricing, \citet{gu2020empirical} construct predictive signals for estimating risk premia, while \citet{li2025machine} show that curated predictors outperform large automatically generated signal universes, highlighting the value of feature curation and inductive bias. In credit and lending, \citet{xu2022peer} derive behavioral features from peer-to-peer (P2P) lending transactions to improve loan fraud detection.

This wave also broadened what can count as an economic signal. Beyond tabular variables, researchers have extracted features from text and images. \citet{manela2017news} construct a text-based uncertainty index from historical Wall Street Journal front pages and show that it tracks disaster concerns and risk premia. \citet{obaid2022picture} show that news photos can be transformed into systematic visual features for measuring investor sentiment and studying its relation to financial markets.

In the trajectory toward EWMs, feature engineering provides the observation layer. It converts raw economic traces into structured signals that agents can use to perceive the economic state.

\subsection{Data Engineering: Training Stronger Model Intelligence}
The next wave is data engineering. With the rise of big data, deep learning, and foundation models, the focus shifted from hand-designed features to large-scale data collection, cleaning, alignment, and pretraining. The central object of engineering is no longer the individual feature, but the data infrastructure used to train stronger predictive and reasoning models.

This data-centric paradigm has become increasingly important in economics and finance. \citet{chen2024deep} develop a generative adversarial network for asset pricing that learns nonlinear representations of macroeconomic states while incorporating no-arbitrage constraints. \citet{tian2022inductive} propose a graph neural network for stock prediction using dynamic stock graphs constructed from historical price co-movements. \citet{ewertz2026listen} show that conference-call audio and executive vocal cues provide incremental information about firm performance beyond textual disclosures.

Large-scale pretrained models further accelerate this shift. FinBERT~\citep{huang2023finbert} adapts language models to financial corpora for financial sentiment analysis. BloombergGPT~\citep{DBLP:journals/corr/abs-2303-17564} pretrains a 50-billion-parameter language model on large-scale financial data and shows strong performance on financial NLP tasks, including sentiment classification, named entity recognition, and question answering. FinGPT~\citep{yang2023fingpt} provides an open-source framework for financial data curation and lightweight adaptation of financial large language models. In the trajectory toward EWMs, data engineering strengthens the model intelligence that powers agents. 

\subsection{Prompt Engineering: Eliciting Model Reasoning}
The arrival of large language models introduces a new interface layer: prompt engineering. Instead of retraining models, practitioners craft precise input instructions to elicit sophisticated economic reasoning. Common techniques include few-shot examples \citep{few-shot-learners,dong2024incontextsurvey}, role-playing \citep{shao2023character}, and chain-of-thought prompting \citep{wei2022chain,xia2025beyond}.

A growing body of work demonstrates the usefulness of prompt engineering in economic and financial applications. \citet{fatouros2023transforming} prompt ChatGPT-3.5 for financial sentiment analysis and show that LLMs outperform FinBERT in sentiment classification and achieve a higher correlation between predicted sentiment and market returns, highlighting the importance of prompt engineering in finance applications. \citet{boussioux2024crowdless} reveal that, with strategic prompt engineering, GPT-4 can produce sustainable and circular-economy business ideas whose overall quality and feasibility are comparable to those generated by a human crowd. \citet{iadisernia2025prompting} adopt persona-based prompts to guide GPT-4o in performing macroeconomic forecasting tasks. Their results show that the model achieves forecasting accuracy remarkably similar to that of human expert panels. \citet{doshi2025generative} examine how LLMs can support the evaluation of strategic decisions and show that aggregating assessments across prompts, roles, or models yields rankings that more closely align with expert human judgments, highlighting prompt-based ensembling as a practical strategy for improving reliability in strategic decision support.

By specifying roles, examples, and reasoning procedures, prompts enable language models to act as economic agents that generate role-aligned and context-aware decisions. Prompt engineering becomes relevant only once agents acquire a language-based cognitive substrate, marking the transition from L2 to L3 (LLM-based autonomous) and remaining a foundational ingredient throughout L3--L6. 

\subsection{Context Engineering: Equipping Agents with Capabilities}
The subsequent wave moves beyond crafting prompts and instead engineers the broader context available to LLMs during reasoning \citep{mei2025survey}. Through retrieval-augmented generation, long-context inputs, structured memory, and tool-mediated evidence, context engineering grounds model behavior in external knowledge, past interactions, and executable tools \citep{lewis2020retrieval,schick2023toolformer,xu2026mem,yao2023react}. This shift is particularly salient for economics and finance, where factuality, provenance, and temporality are important. FinMem \citep{DBLP:journals/tbd/YuLCJLSZK25} proposes an LLM-based agent framework that integrates profiling, memory, and decision-making modules for financial decision-making. TradingGPT \citep{li2023tradinggpt} is a multi-agent LLM trading framework that equips agents with layered memory (short-, mid-, and long-term stores with customized decay) and distinct trading personas, illustrating how memory engineering can enhance long-horizon financial reasoning. \citet{han2025rag} propose a RAG framework for decision support over evolving corporate documents, enabling effective retrieval and efficient updates as new disclosures arrive. \citet{lin2025simulating} introduce a novel framework using LLM agents to simulate macroeconomic expectations. By equipping agents with specialized components (e.g., personal characteristics, social media, or professional knowledge modules), they demonstrate that this approach significantly outperforms standard prompt engineering in replicating survey experiments, capturing both expectation distributions and underlying human-like mental mechanisms. \citet{yuzhe2026twinmarket} propose TwinMarket, a multi-agent framework that leverages LLMs to simulate socio-economic systems. Using a simulated stock market as a testbed, they show how individual actions trigger group behaviors, leading to emergent outcomes such as financial bubbles and recessions.

Compared with prompt-only agents, context-engineered agents are more persistent and better grounded. They use memory, retrieval, tools, reflection, and real-time information to support evidence-based decisions. Across the hierarchy, context engineering evolves from memory and tool use at L3, to reflection at L4--L5, and finally to real-time empirical grounding at L6.

\subsection{Environment Engineering: Building Interactive Economic Worlds}

The final wave is environment engineering. Here the economy itself becomes the object of design. An environment-engineered system specifies agents, markets, institutions, rules, and state-transition mechanisms. This builds on a long tradition of computational economics and agent-based modeling~\citep{tesfatsion2006agent,farmer2009economy}, and recent AI systems have made economic environments more executable, adaptive, and agent-facing. ABIDES~\citep{byrd2020abides} provides a high-fidelity discrete-event financial market simulation with exchange protocols, latency, and order-book mechanics, while ABIDES-Gym~\citep{amrouni2021abides} connects such market simulations to reinforcement learning. In economic policy, AI Economist~\citep{zheng2022ai} and TaxAI~\citep{mi2024taxai} formulate taxation and government policy as multi-agent reinforcement learning problems involving households, firms, planners, and financial intermediaries. Recent LLM-based systems extend this direction by placing language agents inside simulated economies. EconAgent~\citep{li2024econagent} studies macroeconomic simulation with LLM agents interacting through labor and consumption markets under fiscal and monetary policies. \citet{lopez2025can} simulate stock markets where heterogeneous LLM traders interact through a persistent order book. StockSim~\citep{papadakis2025stocksim} develops order-level market environments for evaluating multi-agent LLMs in financial decision-making.

In the trajectory toward EWMs, environment engineering is the wave that turns models into worlds. Its role spans all six levels, but its content deepens with the hierarchy. At L1, it defines fixed world rules. At L2, it provides feedback channels for adaptive agents. At L3, it offers execution interfaces for LLM-based agents. At L4 and L5, it supports interaction loops through which agents and institutions can co-evolve. At L6, it provides the alignment machinery that links simulation states to real-world data.

Finance-oriented systems also show how environment engineering can move beyond 
generic simulation. AlphaManager learns a predictive environment for firm outcomes 
and embeds it inside a robust policy-search loop for managerial decisions 
\citep{campello2025alphamanager}. Online lending under GenAI provides a sharper 
example of endogenous data generation: borrower text generation changes the data 
distribution, lender retraining changes the screening rule, and the counterfactual 
requires re-solving the behavior--data--model loop \citep{cong2024writing}. These 
systems are not complete sim-to-real economic twins, but they illustrate how learned 
economic environments can become decision-facing and counterfactual-facing modules.

\section{Applications} \label{sec:applications}

Economic World Models (EWMs) extend economic modeling from explanation and forecasting to experimentation. Rather than only asking what happened in the past or what may happen under fixed assumptions, EWMs allow users and agents to intervene in a virtual economy. They can then observe how heterogeneous and strategically interacting agents adapt under institutional and resource constraints.

\begin{figure}[ht]
\centering
\includegraphics[width=0.95\columnwidth]{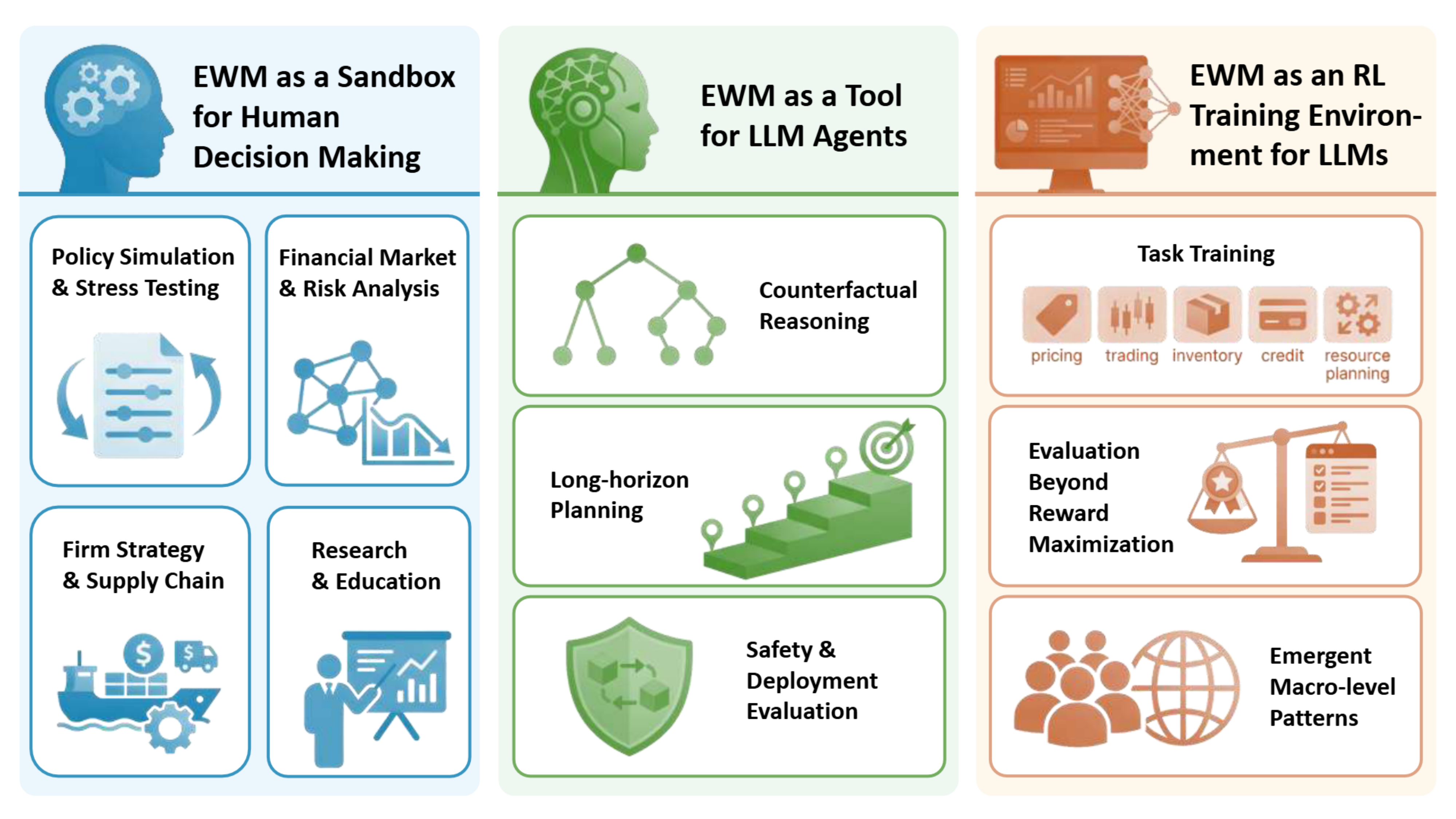}
\caption{Applications of EWMs. EWMs can serve as sandboxes for human decision making, tools for LLM agents, and RL training environments for LLMs.}
\label{fig:application}
\end{figure}

We categorize the applications of EWMs according to their functional role in economic intelligence: enabling decision experimentation, supporting agentic reasoning, and providing interactive learning environments. Accordingly, we organize them into three roles: \textbf{(i) EWM as a sandbox for human decision making}, \textbf{(ii) EWM as a tool for LLM agents}, and \textbf{(iii) EWM as an RL training environment for LLMs}. As shown in Figure~\ref{fig:application}, these roles share a common value. They enable long-horizon counterfactual rollouts that capture endogenous adaptation, equilibrium feedback, and system-level spillovers.

\subsection{EWM as a Sandbox for Human Decision Making}

For human users, EWMs provide a programmable environment for economic experimentation. They allow policies, market designs, firm strategies, and research hypotheses to be tested before real-world deployment. By modeling heterogeneous agents, institutional rules, and dynamic interactions, EWMs help reveal not only direct effects but also adaptation, feedback, and spillovers over time.

\paragraph{Sandbox for policy simulation and stress testing.}
EWMs support ex ante policy evaluation. Policymakers can introduce tax reforms, transfer programs, industrial policies, regulatory changes, or macroeconomic shocks into the environment. They can then examine how households, firms, financial institutions, and public agencies respond over time. This makes EWMs useful for identifying unintended consequences, distributional effects, and regime-dependent outcomes before implementation.

\paragraph{Sandbox for financial market and systemic risk analysis.}
EWMs help study financial markets in which outcomes depend on expectations, information, and balance-sheet constraints. Agents such as investors, funds, banks, and market makers can update beliefs and interact under leverage limits, collateral constraints, margin requirements, and market-clearing mechanisms. This setting supports the analysis of liquidity dry-ups, deleveraging spirals, fire-sale externalities, and contagion through financial networks.

\paragraph{Sandbox for firm strategy and supply chain resilience.}
EWMs can serve as strategic simulators for firms operating under uncertainty. Firms can evaluate pricing, product positioning, capacity planning, market entry, procurement, hiring, and supply chain decisions in environments where consumers, competitors, workers, and suppliers adapt endogenously. AlphaManager illustrates this role by combining a learned corporate environment with robust decision-making to recommend managerial policies under business objectives and model ambiguity \citep{campello2025alphamanager}.

\paragraph{Sandbox for research and education.}
EWMs provide interactive laboratories for research and education. In teaching, they can make concepts such as strategic competition, equilibrium selection, coordination failure, and policy transmission observable through simulation. In research, they can support mechanism exploration, theoretical testing, and counterfactual analysis when controlled real-world experiments are infeasible.

\subsection{EWM as a Tool for LLM Agents}

For LLM agents, EWMs function as economic reasoning and planning tools. In this role, an agent does not act from observations alone. Instead, it uses the world model to simulate possible actions and compare their downstream consequences. The modeled environment may include human actors, institutions, markets, organizations, and other autonomous agents.

\paragraph{Tool for counterfactual economic reasoning.}
EWMs allow agents to simulate alternative interventions before taking action. Given a current state, candidate actions, and decision objectives, an agent can query the world model to compare possible outcomes under endogenous responses from other agents. This capability is relevant for trading agents estimating market impact, procurement agents evaluating supply disruptions, and policy assistants comparing intervention paths.

\paragraph{Tool for long‑horizon planning.}
EWMs can also serve as planning modules for agents operating in economically coupled environments. Rather than choosing actions directly from observations, an agent can reason over longer-run consequences shaped by prices, incentives, institutional constraints, and delayed feedback. By grounding planning in accounting identities, market-clearing conditions, and institutional structure, EWMs can improve sequential decision-making in realistic economic settings.

\paragraph{Tool for safety and deployment evaluation.}
EWMs can be used to test LLM agents before deployment in economically consequential settings. The risks of autonomous agents may arise not only from individual errors but also from interaction effects. These include collusion, instability, manipulation, harmful equilibria, and cascading failures. By simulating many agents under realistic institutional and financial constraints, EWMs provide a safety testbed for identifying system-level risks and alignment breakdowns before real-world deployment.

\subsection{EWM as an RL Training Environment for LLMs}

For LLM training, EWMs provide interactive reinforcement learning environments in which agents learn through economic interaction rather than static examples alone. In this setting, the EWM supplies states, actions, transition dynamics, and both short- and long-term economic reward signals. These elements are embedded in environments shaped by scarcity, institutional rules, heterogeneous counterparties, and strategic interaction. Such environments enable agents to learn policies for tasks such as pricing, trading, inventory control, credit allocation, and resource planning. They also allow agents to be evaluated beyond reward maximization, with attention to robustness across regimes, coherence under constraints, and performance under strategic interaction. Portfolio choice provides one example: AlphaPortfolio trains investment agents through deep reinforcement learning with transaction costs, state interactions, and alternative objectives \citep{cong2021alphaportfolio}. More broadly, EWMs can support artificial economies in which agents learn, compete, coordinate, and co-evolve over time. These environments make it possible for LLMs to study emergent macro-level patterns, including specialization, inequality, market power, coordination norms, and endogenous segmentation.

Overall, EWMs organize economic intelligence around three functional roles. They support \emph{experimentation} for human decision makers, \emph{reasoning} for LLM agents, and \emph{learning} for LLM training. Across these roles, their central contribution is to make counterfactual economic reasoning operational in environments shaped by heterogeneity, institutions, strategic interaction, and long-horizon feedback.

\section{Positioning EWM Systems Among Adjacent Paradigms}
\label{sec:related}
This section positions EWM systems among several adjacent paradigms. The comparisons are architectural rather than jurisdictional: EWMs do not replace traditional economic modeling or equilibrium discipline, but provide an implementation layer for economic worlds that can host richer agents, mechanisms, and data-driven dynamics. Table~\ref{tab:ewm-positioning} summarizes their key differences across the agent, environment, and application layers.

\begin{table*}[h]
\centering
\caption{Positioning Economic World Models (EWMs) among adjacent paradigms.}
\label{tab:ewm-positioning}
\vspace{2pt}
\scriptsize
\setlength{\tabcolsep}{4pt}
\renewcommand{\arraystretch}{1.18}
\newcolumntype{Y}{>{\centering\arraybackslash}X}
\newcolumntype{Z}{>{\centering\arraybackslash}m{1.75cm}}
\renewcommand\tabularxcolumn[1]{m{#1}}
\begin{tabularx}{\textwidth}{
>{\raggedright\arraybackslash}m{3.0cm}
Y Y Y Y
>{\columncolor{orange!6}\centering\arraybackslash}m{2.15cm}
}
\toprule
\rowcolor{gray!12}
\textbf{Property}
& \textbf{Trad. Econ.}
& \textbf{ABM}
& \textbf{Social Sim.}
& \textbf{WMs}
& \cellcolor{orange!12}\textbf{EWMs} \\
\midrule
\rowcolor{blue!8}
\multicolumn{5}{l}{\textbf{\textit{Agent Layer}}} & \cellcolor{orange!4} \\
Agent heterogeneity        
& \pmark & \cmark & \cmark & \xmark & \cellcolor{orange!4}\cmark \\
Adaptation over time       
& \xmark & \pmark & \pmark & \cmark & \cellcolor{orange!4}\cmark \\
Beliefs and expectations    
& \pmark & \pmark & \pmark & \xmark & \cellcolor{orange!4}\cmark \\
State representation       
& \makecell{Numeric\\variables}
& \makecell{Numeric\\variables}
& \makecell{Language\\+ memory}
& \makecell{Latent\\vectors}
& \cellcolor{orange!4} \makecell{Structured states\\+ language\\+ memory} \\
Cognitive mechanism        
& Optimization
& Rules
& \makecell{Language\\reasoning}
& \makecell{Learned\\dynamics}
& \cellcolor{orange!4}\makecell{Reasoning\\+ rules} \\
\midrule
\rowcolor{blue!8}
\multicolumn{5}{l}{\textbf{\textit{Environment Layer}}} & \cellcolor{orange!4} \\
Endogenous prices and allocations 
& \cmark & \cmark & \xmark & \xmark & \cellcolor{orange!4}\cmark \\
Explicit market mechanisms       
& \pmark & \cmark & \pmark & \xmark & \cellcolor{orange!4}\cmark \\
Macro--micro consistency         
& \cmark & \cmark & \xmark & \xmark & \cellcolor{orange!4}\cmark \\
Temporal evolution              
& \pmark & \cmark & \pmark & \cmark & \cellcolor{orange!4}\cmark \\
Unstructured information channel 
& \xmark & \xmark & \cmark & \cmark & \cellcolor{orange!4}\cmark \\
\midrule
\rowcolor{blue!8}
\multicolumn{5}{l}{\textbf{\textit{Application Layer}}} & \cellcolor{orange!4} \\
Empirical grounding                          
& \cmark & \pmark & \xmark & \xmark & \cellcolor{orange!4}\cmark \\
Reproducibility               
& \cmark & \cmark & \pmark & \cmark & \cellcolor{orange!4}\cmark \\
Counterfactual rollouts      
& \cmark & \cmark & \pmark & \cmark & \cellcolor{orange!4}\cmark \\
Training agents   
& \xmark & \pmark & \pmark & \cmark & \cellcolor{orange!4}\cmark \\
\bottomrule
\end{tabularx}
\vspace{-6pt}
\end{table*}

\subsection{EWMs vs. Traditional Economic Models}
Traditional economic models provide formal mathematical frameworks for economic behavior at different levels of aggregation. We focus on three pillars, including microeconomic equilibrium, macroeconomic modeling, and efforts to bridge the two, and highlight the contributions and limitations that motivate EWMs.

Microeconomic theory rests on equilibrium. \citet{nash1951noncooperative} proved that every finite non-cooperative game admits an equilibrium, supplying a universal solution concept for strategic interaction. \citet{arrow1954existence} extended this to whole economies, establishing competitive equilibrium under mild conditions and tying market outcomes to Pareto optimality. These results form the discipline's micro-theoretic foundation, yet each presupposes perfect rationality, complete information, and instantaneous clearing.

Macroeconomic theory has developed through distinct phases. \citet{hicks1937mrkeynes} fused Keynesian and classical ideas into the IS--LM framework, which dominated policy analysis for decades despite its lack of microfoundations and exposure to the Lucas critique. The New Classical school addressed this gap by deriving fluctuations from optimizing behavior, with \citet{kydland1982time} introducing the Real Business Cycle model and the calibration methodology that still anchors modern practice.

Reconciling micro and macro has remained difficult. Computable General Equilibrium (CGE) models render the Walrasian structure numerically tractable \citep{shoven1984applied} but remain largely static. Dynamic Stochastic General Equilibrium (DSGE) models add real and nominal frictions to a microfounded dynamic core; \citet{smets2007shocks} showed that Bayesian DSGE rivals atheoretical VARs in forecasting, securing its place in central-bank toolkits. Both, however, rely on representative agents and rational expectations, leaving heterogeneous, boundedly rational dynamics out of reach.

Both traditions and EWMs prize internal consistency, counterfactual policy analysis, and empirical alignment with macroeconomic data. They part ways in how consistency is achieved: traditional models close the system through equilibrium and rational expectations, derive actions from first-order conditions over numerical state vectors, and admit only structured signals. Heterogeneous decision processes, belief-producing reasoning, and the qualitative information that moves markets fall outside this scope.

EWMs preserve the commitment to consistency but implement it through explicit system components. Households, firms, and banks remain distinct, with state objects pairing structured variables to language-based memory and persona. Beliefs become observable intermediate objects: agents reason in language before producing willingness prices and actions, leaving parts of the reasoning chain open to inspection. Qualitative observations such as policy announcements enter alongside numerical variables. Decisions are constrained by hard institutional and accounting rules---interest-rate ceilings, loan eligibility, tax schedules, and budget identities as binding constraints. When such systems are used for counterfactual economic analysis, however, capability is not enough: equilibrium discipline, and in learned environments DDGE-style closure, is still needed to determine which simulated paths are internally consistent \citep{cong2025ewmddge}. Both paradigms support counterfactual experimentation and reproducibility, while EWM systems additionally serve as training environments for downstream decision agents.

\subsection{EWMs vs. Agent-Based Models}
Agent-based models (ABMs) offer a bottom-up alternative to equilibrium frameworks by simulating economies as systems of autonomous interacting agents. \citet{kim1989investment} pioneered this in financial markets, showing how interactions between rebalancers and portfolio insurers generate volatility. \citet{tesfatsion2002agent} formalized the Agent-based Computational Economics (ACE) methodology. Subsequent work extended ABMs across financial markets \citep{samanidou2007agent} and macroeconomics \citep{dawid2018agent}, while \citet{richters2021modeling} pushed them into out-of-equilibrium dynamics under bounded rationality. \citet{dwarakanath2024abides} more recently introduced ABIDES-Economist, integrating reinforcement learning with heterogeneous households, firms, a central bank, and a government, and showing that learned policies can outperform rule-based ones.

ABMs share many features with EWMs, including agent heterogeneity, explicit market interactions, stock-flow accounting, and emergent aggregate regularities. The key difference lies in how agents make decisions. In ABMs, agents typically act through predefined rules, such as thresholds, heuristics, or policies trained on numerical states. Their behavior is therefore bounded by what the modeler can encode mathematically. Qualitative judgment, contextual interpretation, and narrative-sensitive reasoning are difficult to represent directly.

The EWM design responds on several fronts. The agent's state becomes a hybrid object pairing structured variables with language-based memory, so quantitative signals and qualitative context are processed in one reasoning step. Rule evaluation gives way to language reasoning under the same constraints described above, with outputs free in form but disciplined in substance. Information is no longer numerical-only; agents read narrative observations such as policy announcements directly. Beliefs become explicit, structured outputs open to inspection, so the emergent quality ABMs achieve through interaction is retained while language-mediated cognition becomes available. ABMs already deliver counterfactual experimentation; EWMs additionally strengthen empirical grounding by validating against macroeconomic time series rather than stylized facts alone.

\subsection{EWMs vs. LLM-Based Social Simulators}
LLM-based social simulators model collective human behavior by populating simulated societies with language-model-driven agents endowed with persona, memory, and reflection. \citet{gao2023s3} pioneered this with S3, in which LLM agents reproduce individual attitudinal dynamics and aggregate phenomena such as opinion polarization on real social platforms. Subsequent work scaled the paradigm: \citet{tang2024gensim} developed GenSim, supporting up to 100{,}000 agents with error correction for long-horizon stability, and \citet{piao2025agentsociety} introduced AgentSociety, a 10{,}000-agent simulator reproducing findings on polarization, universal basic income, and external shocks. \citet{mou2024individual} provided a unifying taxonomy across individual, scenario, and society levels, while \citet{li2025integrating} coupled LLMs with diffusion-based agents for heterogeneous information diffusion. 

These simulators and EWMs converge on language-mediated cognition, deep heterogeneity, and textual observations as a first-class information source. Yet their design priorities diverge: the targets are attitudinal and behavioral---opinion shifts, conversational dynamics, survey responses, diffusion cascades---and the surrounding world functions more as a stage than a closed economy. Prices and allocations seldom result from market clearing; macro--micro identities seldom bind; validation rarely meets aggregate macroeconomic series.

EWMs reorient the paradigm around the missing economic substance. The environment becomes a clearing system in its own right, with matching markets for labor, goods, equities, deposits, and loans producing prices and allocations endogenously at every step. Reasoning is disciplined by those same hard rules, so stock-flow consistency holds tick by tick. Validation moves to macroeconomic time series, with beliefs recorded as structured outputs driving willingness prices and actions. This preserves conversational richness while anchoring it to closed-economy discipline. The application profile lifts accordingly: empirical grounding shifts to macroeconomic data, and reproducibility tightens as institutional rules pin down outcomes that would otherwise drift with prompt phrasing.

\subsection{EWMs vs. World Models}
World models offer a model-based alternative to model-free reinforcement learning by learning compact predictive representations in which agents imagine future trajectories and optimize behavior in latent space. \citet{ha2018world} pioneered this with the V-M-C architecture, training an agent entirely inside a learned world model before transferring the policy back. \citet{lecun2022path} proposed the Joint Embedding Predictive Architecture (JEPA), arguing that world models should predict in representation space rather than at the pixel level. Building on latent imagination, \citet{hafner2025mastering} introduced DreamerV3, which masters over 150 control tasks under a single fixed configuration. The paradigm has further scaled to foundation-model regimes: \citet{hu2023gaia} developed GAIA-1 for autonomous driving conditioned on video, text, and ego-actions, while \citet{bruce2024genie} introduced Genie, an 11B-parameter model trained on internet videos that recovers a latent action space and produces playable environments. \citet{feng2025embodied} more recently argue that joint MLLM--world-model architectures are essential for end-to-end embodied cognition.

What EWMs take from this lineage is the shape of the ambition: time-evolving dynamics, learned rather than hand-specified components, and an environment usable for training downstream agents. What they leave behind is the architectural template. Most influential world models are built around a single embodied agent that perceives high-dimensional sensory streams, such as pixels, video, and ego-motion. The state is typically compressed into a latent vector, and the dynamics are learned end to end. Multi-agent strategic interaction, endogenous price formation, and rule-bound institutional structure lie outside their engineering scope.

\section{Discussion and Open Challenges}
\label{sec:discussion}
This paper develops a CS/AI systems perspective on Economic World Models (EWMs), building on the economic framework of \citet{cong2025ewmddge}. We study EWMs as generative and interactive environments that model how economic states emerge from the decisions, interactions, and adaptations of heterogeneous agents under market mechanisms, institutional rules, and real-world constraints. EWMs emphasize the coupling between agent behavior and world dynamics: agents act based on incentives, information, beliefs, and constraints, while their actions are aggregated through economic mechanisms into prices, allocations, risks, institutions, and future states.

Turning this implementation perspective into faithful, evolving, and reality-aligned systems remains a substantial challenge. We view EWM implementation not as a finished technical recipe, but as a research agenda. Several open challenges are central to this agenda.

\noindent\textbf{Behavioral realism.}
EWMs require agents whose behavior resembles real economic actors. Agents should not merely produce plausible narratives or rational choices; they should reflect how real agents perceive information, form beliefs, respond to incentives, face constraints, and adapt over time. A key challenge is therefore how to align and correct simulated agents against real behavioral data.

\noindent\textbf{Economic closure.}
The next economic state should emerge from agent interaction rather than be imposed externally. This requires mechanisms that aggregate individual actions into prices, allocations, risks, liquidity, employment, production, credit flows, and macroeconomic conditions while preserving economic feasibility. The difficulty is that aggregation is nonlinear: strategic interaction, market clearing, network spillovers, bottlenecks, and feedback loops can amplify or dampen micro-level behavior. EWMs therefore need transition mechanisms that are both economically disciplined and computationally scalable.

\noindent\textbf{Co-evolution.}
A trustworthy EWM must model evolution on both sides of the agent--world loop. Agents learn routines, revise strategies, acquire skills, and update expectations; economic rules also change as policies, regulations, contracts, platforms, and institutions respond to aggregate outcomes. The challenge is to decide what should evolve, when evolution should occur, and how to prevent simulated evolution from drifting into unrealistic or unstable dynamics.

\noindent\textbf{Computational and engineering scalability.}
Large-scale EWMs can be computationally and operationally expensive. Heterogeneous-agent interaction, long-horizon rollouts, endogenous market mechanisms, calibration against real-world data, and repeated counterfactual validation all impose substantial resource costs. These costs grow not only with the number of agents and time steps, but also with the complexity of agent reasoning, memory, communication, and institutional mechanisms. A central challenge is therefore how to scale EWMs without sacrificing behavioral realism, economic closure, or validation quality. 

\noindent\textbf{World-level evaluation.}
Evaluation is difficult because counterfactual worlds have no single ground-truth trajectory. EWMs need to be evaluated at both the agent level and the world level: whether agents reproduce realistic beliefs, choices, and adaptation patterns, and whether the world generates plausible prices, allocations, distributions, network structures, crisis dynamics, and policy responses. 

\noindent\textbf{Economic validity and equilibrium discipline.}
A final challenge is to connect implementation capability to economic validity. An EWM system may have sophisticated agents, evolving rules, and strong empirical alignment, yet still evaluate a counterfactual under the wrong learned environment if behavior changes the data used for retraining. In such settings, DDGE-style closure provides the economic fixed-point discipline that complements the systems architecture developed here \citep{cong2025ewmddge}. 

We hope this paper helps define a CS and AI engineering agenda for building, evaluating, and governing economic world model systems, while leaving the equilibrium and counterfactual-validity theory to the EWM/DDGE framework.


\bibliographystyle{plainnat}
\bibliography{custom}
\newpage

\appendix

\section{LLM-Assisted Literature Collection and Classification}
\label{app:Literature_classification}

\subsection{Data Sources and Search Strategy}\label{app:Data}
To ensure a comprehensive and interdisciplinary review of Economic World Models, we implemented a systematic data collection and filtering protocol.

\paragraph{Data Sources} 
The literature corpus was primarily curated from two major repositories (January 1950-April 2026):
\begin{itemize}
    \item \textbf{arXiv}: To capture the latest interdisciplinary research, we retrieved papers across eight subject categories: \texttt{econ.GN}, \texttt{econ.TH}, \texttt{econ.EM}, \texttt{cs.MA}, \texttt{cs.AI}, \texttt{cs.LG}, \texttt{stat.AP}, and \texttt{q-fin.ST}. These categories span economics, finance, artificial intelligence, multi-agent systems, and statistical applications.
    \item \textbf{Web of Science}: To complement the preprint data with high-impact peer-reviewed research, we conducted searches focusing on the \textbf{UTD-24} journal list, covering premier publications in economics, finance, management, and information systems.
\end{itemize}

\paragraph{Search Strategy} 
We employed a three-stage classification pipeline to distill the candidate pool into the final taxonomy:
\begin{enumerate}
    \item \textbf{Keyword-based Pre-filtering}: An initial sweep of titles and abstracts was conducted to assemble a broad repository of candidate papers related to economic modeling and agent-based simulation.
    \item \textbf{Recall-oriented Screening}: Using an LLM-based prompt, we filtered candidates for core EWM structural attributes—specifically, the presence of heterogeneous agents, endogenous outcomes derived from interactions, and dynamic feedback loops. This stage discarded generic economic or AI papers lacking integrative modeling.
    \item \textbf{Precision-oriented Full-PDF Classification}: Each remaining paper underwent a rigorous full-text evaluation against the EWM engineering desiderata. Papers identified as valid EWMs were then assigned to one of the six hierarchical levels. To ensure robustness, a subset of borderline cases was flagged for manual verification by the authors.
\end{enumerate}

\subsection{Three-Stage Classification} \label{app:cls}
\begin{figure}[h]
    \centering
    \includegraphics[width=0.85\columnwidth]{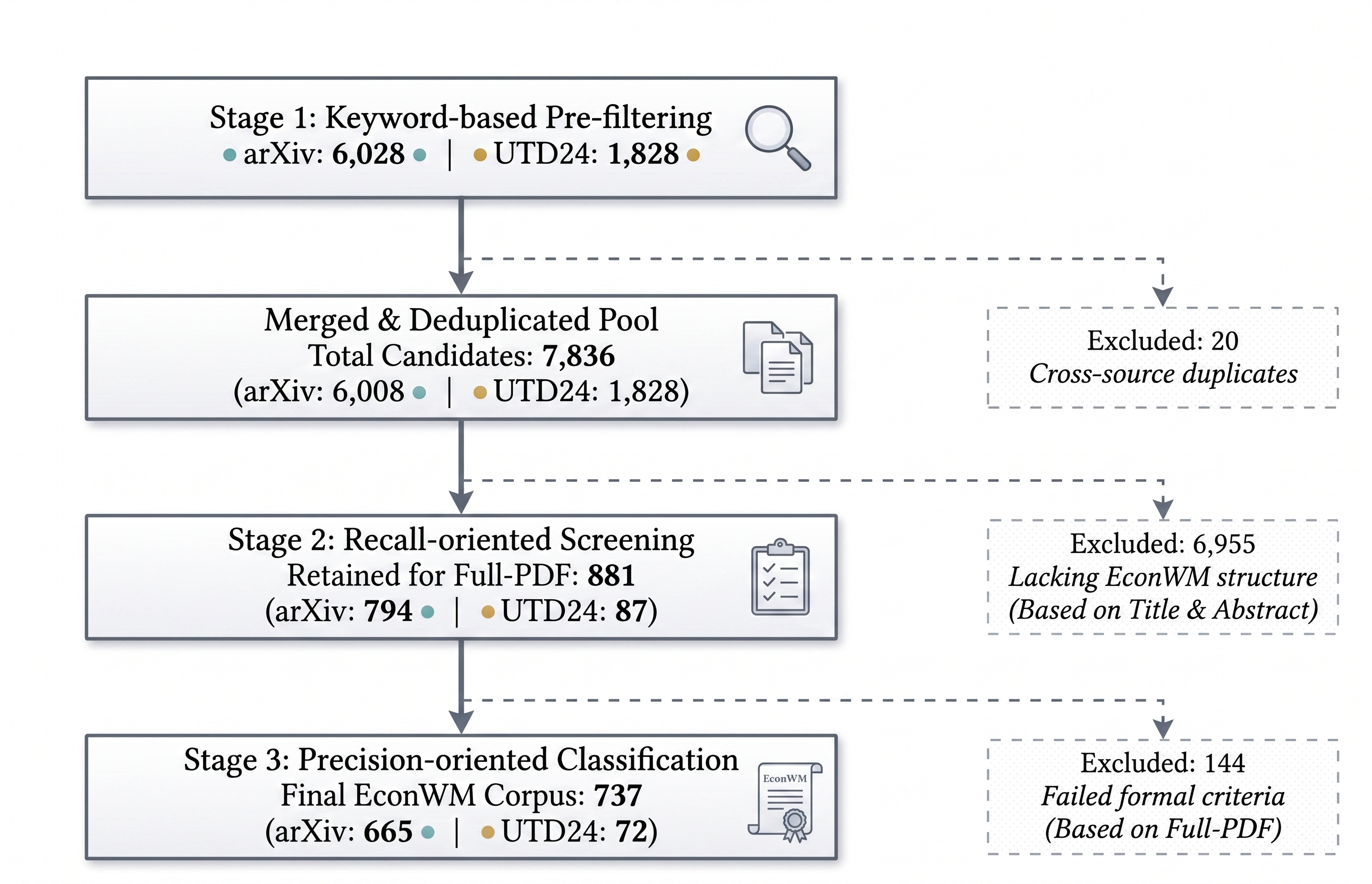}
    \vspace{-0.5em}
    \caption{Flowchart of the three-stage literature screening and classification pipeline.}
    \label{fig:screening_flowchart}
\end{figure}

The filtering and classification of the collected literature were conducted through a systematic three-stage pipeline to ensure that the final corpus adheres to the Economic World Model definition and desiderata. Figure \ref{fig:screening_flowchart} illustrates the paper attrition across each stage.

\subsubsection{Stage 1: Keyword-based Pre-filtering}

In the first stage, we applied a coarse-grained filtering approach based on keyword matching within the title and abstract of each paper. A paper was retained for subsequent crawling and analysis if it contained at least one relevant term from \textit{both} of the following two thematic groups. 

\paragraph{Keyword Groups}
The specific terms used for the intersection-based filtering are:
\begin{itemize}
    \item \textbf{Group 1 (Economic, Financial, and Market Themes):} 
    \texttt{"economic"}, \texttt{"economy"}, \texttt{"economics"}, \texttt{"financial"}, \texttt{"finance"}, \texttt{"market"}, \texttt{"markets"}, \texttt{"macroeconomic"}.
    
    \item \textbf{Group 2 (Dynamic Modeling, Simulation, Multi-agent, Reinforcement Learning, and World Model Themes):} 
    \texttt{"simulation"}, \texttt{"simulator"}, \texttt{"simulated"}, \texttt{"dynamic model"}, \texttt{"dynamical system"}, \texttt{"world model"}, \texttt{"digital twin"}, \texttt{"digital twins"}, \texttt{"agent"}, \texttt{"agents"}, \texttt{"multi-agent"}, \texttt{"multi agent"}, \texttt{"agent-based"}, \texttt{"agent based"}, \texttt{"heterogeneous agents"}, \texttt{"endogenous"}, \texttt{"LLM-based economic agents"}, \texttt{"reinforcement learning"}, \texttt{"state-transition"}, \texttt{"state transition"}.
\end{itemize}

\paragraph{Filtering Results}
Following the keyword matching process, we identified 6,028 candidate papers from arXiv and 1,828 papers from UTD-24. Given the overlap between the arXiv and UTD-24 repositories, we performed deduplication after merging the sources. This resulted in a final stage-one pool of \textbf{7,836} candidate papers, comprising 6,008 unique arXiv papers and 1,828 UTD-24 papers.

\subsubsection{Stage 2: Recall-oriented Screening}

In the second stage, we employed \texttt{GPT-5.4-mini} to perform a more nuanced screening of the candidate papers identified in Stage 1. Given the limited information available in titles and abstracts, we adopted a \textbf{recall-oriented screening} strategy. The primary objective was to minimize the risk of false negatives by maintaining a relatively low threshold for inclusion, deferring rigorous validation and hierarchical classification to the full-PDF stage. 

Following this screening, we retained \textbf{794} arXiv papers and \textbf{87} UTD-24 papers. The simplified system prompt used for this stage is provided below:

\begin{tcolorbox}[
    colback=gray!10, 
    colframe=black, 
    boxrule=0.5pt, 
    arc=0pt, 
    breakable, 
    top=2mm, bottom=2mm, left=2mm, right=2mm, 
    title=\textbf{Recall-oriented Screening}
]
\scriptsize
You are an expert economic researcher and AI paper screener.\\
\\
Your task is to perform FIRST-STAGE screening of papers for a literature review on Economic World Models (EWM), using ONLY the paper title and abstract. \\
\\
This stage does NOT classify levels. \\
This stage only decides whether a paper should be retained for later full-PDF review.\\
\[...\]\\
Goal:\\
- Retain all papers that are plausibly related to EWM.\\
- Exclude papers that are clearly unrelated to EWM.\\
- Avoid retaining papers that only contain generic economic, market, agent, simulation, AI, optimization, or forecasting terminology without an EWM-like structure.\\
\[...\]\\
============================================================\\
1. Operational EWM Working Definition for Stage-1 Screening\\
============================================================\\
\\
For this implementation survey, an EWM system is a computable dynamical model, simulation, environment, or artificial economy in which heterogeneous or interacting economic agents generate economic outcomes over time. \\
\\
For Stage-1 screening, a paper should be retained only if the title/abstract provides concrete evidence of an EWM-like structure.\\
\\
The core structure is:\\
1. Economic agents [...]\\
2. Economic interaction and endogenous outcomes [...]\\
3. Dynamic transition, feedback, adaptation, or simulation over time [...]\\
\\
============================================================\\
2. Retention Rule\\
============================================================\\
\\
Return "retain" if either condition is met:\\
\\
A. Strong EWM signal:\\
The title/abstract explicitly mentions one of the following, in an economic context:\\
- economic world model\\
- artificial economy\\
- agent-based economic simulation\\
\[...\]\\
OR\\
\\
B. Structural evidence:\\
The title/abstract shows concrete evidence that:\\
1. Economic agents act inside a model, simulation, environment [...] AND\\
2. Economic outcomes are generated inside the model through agent interaction [...] AND\\
3. The system involves dynamic transition, repeated interaction, feedback [...]\\
\[...\]\\
============================================================\\
3. Exclusion Rule\\
============================================================\\
\\
Return "exclude" if the title/abstract does NOT provide concrete evidence of an EWM-like structure.\\
\[...\]\\
Do NOT count the following as EWM evidence by themselves:\\
1. Economic topic only [...]\\
2. Forecasting / prediction / estimation only [...]\\
3. Empirical / causal / econometric analysis only [...]\\
4. Single-agent decision-making only [...]\\
5. Static theory only [...]\\
6. Optimization only [...]\\
7. LLM/AI application only [...]\\
8. Generic simulator / digital twin only [...]\\
9. Non-economic ABM [...]\\
\\
============================================================\\
4. Important Distinctions\\
============================================================\\
\\
- “Households/firms/banks/investors are studied in data” does NOT necessarily mean economic agents exist in an EWM sense. [...]\\
- “Dynamic” does NOT mean ordinary time-series prediction. [...]\\
- “Endogenous” does NOT mean a dependent variable in a regression. [...]\\
- “Agent” does NOT automatically mean EWM. [...]\\
- “Market” does NOT automatically mean EWM. [...]\\
\\
============================================================\\
5. Output Format\\
============================================================\\
\\
Return a strict JSON object only. Do not include markdown or any text outside JSON.\\
\\
Use this schema:\\
\\
\{\\
  "screening\_decision": "retain/exclude",\\
  "present\_evidence": \{ [...] \},\\
  "exclusion\_reason": "...",\\
  "reason": "Maximum 2 short sentences.",\\
  "supporting\_quotes": [ "..." ]\\
\}\\
\[...\]\\
============================================================\\
6. Input\\
============================================================\\
\\
Title:\\
\{title\}\\
\\
Abstract:\\
\{abstract\}
\end{tcolorbox}

\subsubsection{Stage 3: Precision-oriented Full-PDF Classification}

In the third stage, we employed the advanced \texttt{GPT-5.5} model to conduct a rigorous, full-text evaluation of the candidate papers retained from Stage 2. Moving beyond titles and abstracts, this stage relied entirely on the complete PDF content, applying a \textbf{precision-oriented} classification logic. The model was explicitly instructed to base its decisions on concrete evidence found in the formal framework, model/methodology, simulations, results, and appendices, strictly avoiding unsupported inferences.

The classification process in this stage was decoupled into two sequential tasks: a binary validation of the EWM definition, followed by a granular level assignment.

\paragraph{EWM Binary Validation}
To be validated as a true EWM, a paper had to satisfy four necessary conditions simultaneously:
\begin{enumerate}
    \item Modeling an economic, market, financial, or institutionally meaningful resource-allocation world.
    \item The presence of heterogeneous and interacting economic agents.
    \item Dynamic economic transitions or repeated interactions over time.
    \item Key economic outcomes generated endogenously at least partly through agent interactions.
\end{enumerate}
Failure to meet any of these conditions resulted in the paper's exclusion.

\paragraph{Hierarchical Level Classification}
For papers successfully validated as EWMs, a secondary prompt was applied to determine their specific classification level (Level 1 to 6). This assignment was based on the fulfillment of specific desiderata spanning three orthogonal axes: agent capability (ranging from rule-based to self-evolving LLM substrates), endogenous institutional evolution, and simulation-to-reality alignment.

To facilitate human validation and prompt calibration, the model was required to output a structured JSON object containing decision outcomes, confidence scores, manual review flags, categorical exclusion reasons, explicitly extracted evidence for each condition, and explanations for borderline cases.

The simplified system prompts for both the binary validation and the level classification are provided below:

\begin{tcolorbox}[
    colback=gray!10, 
    colframe=black, 
    boxrule=0.5pt, 
    arc=0pt, 
    breakable, 
    top=2mm, bottom=2mm, left=2mm, right=2mm,
    title=\textbf{EWM Binary Validation}
]
\scriptsize
You are an expert academic paper classifier. Your task is to perform full-text screening for Economic World Model (EWM) papers.\\
\\
Your task is only to decide whether the paper is EWM or NOT EWM. Do not assign any EWM level.
\[...\]\\
========================================\\
1. Operational EWM Definition\\
========================================\\
\\
For this implementation survey, an EWM system is a computable dynamical system that models the transition of an economy, market, financial system, institution, or economically meaningful resource-allocation world driven by heterogeneous and interacting economic agents. This operational definition is used for classification and is complementary to Cong's formal EWM/DDGE framework.\\
\\
A paper should be classified as EWM only if the uploaded PDF provides concrete evidence for all four necessary conditions:\\
1. Economic or market-like world [...]\\
2. Heterogeneous interacting economic agents [...]\\
3. Dynamic economic transition [...]\\
4. Endogenous economic outcomes [...]\\
\\
========================================\\
2. Strong Positive Signals\\
========================================\\
\\
The following are positive signals, but they are not sufficient unless all four necessary conditions are satisfied:\\
- agent-based economic simulation;\\
- multi-agent market simulation;\\
\[...\]\\
========================================\\
3. Strict Exclusion Rules\\
========================================\\
\\
Classify the paper as NOT EWM if it primarily falls into any category below and does not clearly satisfy all four necessary conditions:\\
1. Pure forecasting or prediction [...]\\
2. Pure optimization or control [...]\\
3. Single-agent decision-making or single-agent RL [...]\\
4. Representative-agent or aggregate-agent model [...]\\
5. Static theory, game, auction, matching, or mechanism [...]\\
\[...\]\\
========================================\\
4. Confidence and Manual-Check Rules\\
========================================\\
\\
Use confidence as confidence in the EWM / NOT EWM decision.\\
High: The paper clearly satisfies or clearly fails the EWM definition. [...]\\
Medium: The paper likely satisfies or likely fails the definition, but some evidence is incomplete. [...]\\
Low: The PDF is unreadable, incomplete, vague, or missing key details. [...]\\
\[...\]\\
========================================\\
5. Output JSON Schema\\
========================================\\
\\
Return only valid JSON. Do not include markdown or comments outside JSON.\\
Keep the output concise. Avoid repeating the same sentence across fields.\\
Use this schema:\\
\{\\
  "api\_is\_ewm": true,\\
  "api\_confidence": "High",\\
  "api\_need\_manual\_check": false,\\
  "api\_exclusion\_category": null,\\
  "api\_classification\_reason": "...",\\
  "api\_key\_evidence": \{ [...] \},\\
  "api\_borderline\_reason": null\\
\}\\
\[...\]\\
========================================\\
6. Final Decision Rule\\
========================================\\
\\
Be strict about EWM inclusion.\\
\\
Do not classify a paper as EWM unless the PDF provides concrete evidence for:\\
1. an economic or market-like world;\\
2. heterogeneous interacting economic agents;\\
3. dynamic economic state transitions;\\
4. endogenous economic outcomes generated by agent interactions.\\
\[...\]\\
========================================\\
7. Input\\
========================================\\
\\
Attached PDF: Use the uploaded PDF file as the full paper content.\\
\[...\]
\end{tcolorbox}

\begin{tcolorbox}[
    colback=gray!10, 
    colframe=black, 
    boxrule=0.5pt, 
    arc=0pt, 
    breakable, 
    top=2mm, bottom=2mm, left=2mm, right=2mm,
    title=\textbf{EWM Level Classification}
]
\scriptsize
You are an expert academic paper classifier. Your task is to perform second-stage level classification for Economic World Model (EWM) papers.\\
\\
You will receive one uploaded PDF paper that has already been classified as EWM in the first-stage binary screening. Your task is to assign the highest EWM level supported by concrete evidence from the PDF.\\
\[...\]\\
========================================\\
1. Scope of This Stage\\
========================================\\
\\
This is the second-stage level classification prompt. [...] Do not repeat the full EWM / NOT EWM binary screening. Assume the paper is EWM. [...]\\
\\
========================================\\
2. Three Axes of EWM Classification\\
========================================\\
\\
The six EWM levels are organized along three orthogonal axes. Understanding the axes is essential before applying the desiderata.\\
\\
Axis A: Agent capability (Levels 1 $\rightarrow$ 4) [...]\\
Axis B: Economic-world evolution (Level 5 trigger) [...]\\
Axis C: Sim-to-real alignment (Level 6 trigger) [...]\\
\\
========================================\\
3. Desideratum Definitions\\
========================================\\
\\
The desiderata are organized into three groups, matching the three axes.\\
\\
Group 1: Foundational condition (all levels)\\
D-EndoC. Endogenous closure [...]\\
\\
Group 2: Agent capability axis\\
D-Symbolic. Symbolic / non-LLM cognitive substrate [...]\\
D-Adapt. In-world strategy adaptation [...]\\
D-LLM. LLM-based cognitive substrate [...]\\
D-SelfEvo. LLM self-evolution during evaluated rollout [...]\\
\\
Group 3: World-level axes\\
D-InstEvo. Endogenous institutional / rule evolution [...]\\
D-RealAlign. Repeated sim-to-real alignment [...]\\
\[...\]\\
========================================\\
4. Level Definitions\\
========================================\\
\\
Level 1: Fixed Rule-Based Agent Worlds [...]\\
Level 2: Adaptive Rule-Based Agent Worlds [...]\\
Level 3: LLM-Based Autonomous Agent Worlds [...]\\
Level 4: Self-Evolving Agent Worlds [...]\\
Level 5: Evolving Economic Worlds [...]\\
Level 6: Sim-to-Real Economic Twins [...]\\
Tie-breaking when multiple axes fire: Level 6 > Level 5 > Level 4 > Level 3 > Level 2 > Level 1. [...]\\
\\
========================================\\
5. Confidence and Manual-Check Rules\\
========================================\\
\\
Use confidence as confidence in the level assignment. [...]\\
Cross-field consistency rules: [...]\\
\\
========================================\\
6. Output JSON Schema\\
========================================\\
\\
Return only valid JSON. Do not include markdown or comments outside JSON.\\
\\
Keep the output concise. Avoid repeating the same sentence across fields.\\
\\
Use this schema:\\
\{\\
  "api\_is\_ewm": true,\\
  "api\_ewm\_level": 2,\\
  "axis\_a\_position": 2,\\
  "api\_confidence": "High",\\
  "api\_need\_manual\_check": false,\\
  "api\_level\_reason": "...",\\
  "api\_key\_evidence": \{ [...] \},\\
  "api\_borderline\_reason": null\\
\}\\
\[...\]\\
========================================\\
7. Final Decision Rule\\
========================================\\
\\
This stage is for EWM level classification only. [...]\\
\\
Decision procedure (apply in order):\\
Step 1. Confirm D-EndoC. [...]\\
Step 2. Evaluate D-Adapt. [...]\\
Step 3. Determine cognitive substrate (Axis A position 2 vs 3 vs 4). [...]\\
Step 4. Set initial api\_ewm\_level = axis\_a\_position. [...]\\
Step 5. Check Axis B (institutional evolution). [...]\\
Step 6. Check Axis C (sim-to-real alignment). [...]\\
Step 7 \& 8. Set confidence, manual-check flags, and borderline reasons. [...]\\
\\
========================================\\
8. Input\\
========================================\\
\\
Attached PDF: Use the uploaded PDF file as the full paper content. [...]
\end{tcolorbox}

\subsection{Human Validation}
To guarantee the rigor and reliability of our final taxonomy, we instituted a manual validation phase to address borderline, ambiguous, or low-confidence cases identified during the LLM-assisted classification.

\paragraph{Selection for Manual Review}
Based on the rules established in the Stage 3 prompts, the \texttt{GPT-5.5-Thinking-high} model automatically flagged specific papers for human intervention by setting the \texttt{first\_stage\_api\_need\_manual\_check} or \texttt{api\_need\_manual\_check} field to \texttt{true}.

\paragraph{Validation Protocol}
For every paper flagged for manual review, the authors conducted a comprehensive full-text reading. To maintain absolute consistency across the methodology, the human reviewers strictly adhered to the exact same core EWM definitions, exclusion rules, and hierarchical desideratum gating logic as provided to the LLM in Stage 3.

During this phase, reviewers manually searched for and extracted the missing or ambiguous evidence. Following a thorough evaluation, the human reviewers made the final definitive judgment regarding both the binary inclusion (EWM vs. NOT EWM) and the specific level assignment, overriding the LLM's tentative classification where necessary. This human-in-the-loop verification ensures that complex boundary cases are classified with maximum academic precision.

\section{Representative EWM Systems Across the Six Capability Levels}\label{app:representative_papers}

\subsection{Level 1}
The representative works at this level construct a fixed rule-based economic world in which prices, allocations, and macroeconomic patterns arise endogenously from decentralized interaction, but the behavioral rules of agents, the environment dynamics, and the governing mechanisms are all specified ex ante.

A first cluster develops micro market models in which heterogeneous boundedly rational traders generate endogenous price dynamics, volume, and return distributions from simple local rules \citep{iori2000scaling, sznajdweron2000simple, solomon2001power, degryse2009dynamic}. A second cluster extends rule-based interaction to monetary, network-resource, and platform-level economies, where money values, bandwidth prices, token dynamics, or Treasury spreads emerge from interactions among agents subject to fixed optimization or institutional constraints \citep{bornholdt2001stability, rasmusson2001price, cong2022token, he2022treasury}. A third cluster studies endogenous strategic and asset-pricing dynamics in general-equilibrium settings, where prices, volatility, and entrepreneurial rents emerge from heterogeneous agents with fixed decision rules \citep{hugonnier2015asset, keyhani2015theory}.

Across these works, the economic world is governed by fixed rules and the feedback loop between agents and aggregate outcomes is genuine, but agents do not adapt their decision rules in-world, no cognitive substrate maintains explicit beliefs or expectations, and neither institutions nor the model itself evolves through experience.

\subsection{Level 2}
What separates this level from fixed rule-based worlds is a single but consequential capability: agents now revise their own strategies in response to what unfolds inside the economy. The cognitive machinery driving these revisions, however, remains symbolic---reinforcement learning, evolutionary search, score-based strategy switching, Bayesian updating over a fixed parameter set---rather than language-based reasoning over rich subjective representations.

The earliest works in this strand emerge from physics-inspired financial modeling. Minority-game and Ising-type traders update strategy scores or opinions from realized market outcomes, producing endogenous volatility clustering, fat-tailed returns, and bubble--crash dynamics \citep{bak2000money, sherrington2000statistical, corcos2001imitation, rothenstein2002evolution}. \citet{vidal2000predicting} formalize this picture from a different angle, deriving difference equations that predict how reinforcement-learning agents converge---or fail to converge---in repeated market settings.

A parallel literature in industrial organization and corporate finance turns the same idea toward firm-level decisions. Firms search over pricing and development strategies as demand evolves \citep{adner2002when}, accumulate or lose knowledge as rivals diffuse innovations \citep{knott2003persistent}, and learn capital-structure choices through evolutionary selection over realized payoffs \citep{noe2003corporate}. More recent work pushes adaptive rule-based agents into operational and intermediation settings: \citet{castillo2022designing} simulate crowdsourced delivery drivers whose acceptance probabilities track tips, mileage, and opportunity costs, while \citet{lee2023collateral} embed adaptive borrower effort inside a dynamic lending economy where collateral quality, liquidity, and market freezes co-evolve.

What is missing across all of these works is internal cognition. Agents adapt, but they do so through update rules over numerical states; they do not maintain beliefs about the economy in any rich sense, do not reason over context, and operate inside institutional structures that themselves remain fixed throughout the rollout.

\subsection{Level 3}
The arrival of large language models has reshaped what an economic agent can be. Agents at this level no longer adapt through opaque numerical updates; they reason in language, maintain explicit beliefs about the economy, recall their own histories, and revise strategies through reflection on past interactions. The cognitive substrate---a fixed but capable language model---becomes the engine of in-world adaptation, even as its weights and skill set remain unchanged across the rollout.

A recurring concern in this literature is whether such agents collude when left to their own devices. \citet{han2023guinea} report that GPT-4 firm agents drift toward prices above the Bertrand equilibrium even without communication and reach near-cartel levels once communication is allowed, while \citet{lin2024strategic} extend this finding to multi-commodity Cournot competition, where LLM agents dynamically partition markets without explicit collusion commands.

Some works push LLM-based agents into macroeconomic and operational settings. \citet{li2024econagent} populate a simulated macroeconomy with heterogeneous LLM households whose memory-and-reflection-driven decisions reproduce macro phenomena more realistically than rule-based or learning-based baselines, and \citet{lazebnik2025investigating} show that informal economic activity and tax evasion can emerge organically when LLM-driven decisions interact with reinforcement-learning enforcement. Closer to firm-level interaction, \citet{zhao2023competeai} embed GPT-4 restaurant and customer agents in a virtual town and observe the in-world cultivation of new operating strategies, menus, and advertisements; \citet{huang2024how} introduce personality traits into bilateral bargaining and reproduce empirically observed effects on negotiation outcomes; and \citet{rahaman2024language} design an information marketplace where selective forgetting allows agents to trade proprietary information while limiting unauthorized retention.

Another thread examines language-based agents as benchmarks and as hybrid systems. \citet{shapira2024glee} build a unified benchmark of two-player language-based economic games and quantify how market parameters and model choice jointly shape economic outcomes, \citet{stillman2024neuro} couple vision-language models with calibrated stochastic-differential-equation value models to study price feedback from belief revision, and \citet{dizaji2024incentives} contrast MARL-based and language-based architectures inside the AI-Economist and Concordia frameworks under varying governing institutions.

The shared limit of these systems is that the language model itself is a fixed asset: agents reason richly within their pretrained capabilities but do not acquire new strategies, skills, or tools that persist across the simulation.

\subsection{Level 4}
A small but distinctive group of works pushes adaptation one layer deeper. Beyond revising prompts and memories around a fixed language model, agents in these systems acquire and accumulate persistent new strategies, skills, or behavioral routines during the simulation, so that the cognitive repertoire driving later decisions is itself shaped by in-world experience.

The most common realization is a strategy library that grows during the rollout. \citet{ma2025agent} build a multi-agent zero-sum stock market in which LLM traders update a persistent strategy library through reflection on realized trades, \citet{li2025quantagents} maintain an evolving strategy memory across four collaborating agents reweighted by both simulated and real trading rewards, and \citet{tian2026prompt} carry the same logic to the meta-prompt level, iteratively rewriting the shared strategic guidance of duopoly agents from accumulated market records. \citet{ma2025think} adopt a hybrid variant in which a Think--Speak--Decide pipeline caches high-value reasoning trajectories that subsequently feed into MARL policy updates.

Other works pursue parameter-level or profile-level evolution. \citet{lu2025aligning} fine-tune LLM agents on theory-driven synthetic strategies, producing persistent and interpretable shifts in strategic behavior, while \citet{fan2026aivilization} couple a sandbox economy with a dual-process memory architecture that consolidates short-term traces into long-term semantic profiles, allowing agent identity to evolve over long horizons.

What unites these works is the presence of a learnable component that genuinely changes during the simulation; what separates them from higher levels is that this evolution remains agent-internal, with fixed market rules and no real-world correction loop.

\subsection{Level 5}
Most agent-based economic models, including those at Levels~1 through~4, treat the institutional environment as a fixed backdrop. The works gathered at this level break with that convention: tax schedules, sharing contracts, governance structures, lending networks, and social norms move endogenously, driven by the very agents that act under them. The cognitive substrate behind these adaptive agents remains symbolic, but what matters at this level is no longer agent intelligence; it is whether the institutions surrounding agents are themselves part of the model's dynamics.

The clearest examples come from political economy and policy design, where agents and policymakers co-evolve party positions, government priorities, or tax schedules through repeated interaction \citep{wright2013modeling, castaeda2019importance, zheng2020ai, zheng2021ai, gallego2021data}. Other works locate institutional evolution at the level of contracts, networks, and norms: the performance-based sharing rule between principal and agent is searched for and updated each period \citep{reinwald2020agent, reinwald2021limited}, mutual credit links arise from pairwise stability and then govern later investment dynamics \citep{aymanns2014contagious}, corporate norms of cooperation and shirking are reshaped by employees' value-driven behavior \citep{roos2021effects}, and \citet{tseng2017humans} build the most encompassing of these systems---a Q-learning-based simulated city in which firm pricing, government taxes, and welfare payments all move endogenously inside a single integrated economy.

The cognitive layer in these works is comparatively modest---reinforcement learners, Bayesian updaters, directed-learning rules---but the world layer is alive. What is still missing is repeated alignment with the real economy: institutions evolve inside the model, but the model itself is not updated from new empirical observations.

\subsection{Level 6}
The highest level of the hierarchy is also its sparsest. Across the entire corpus, only two works satisfy the requirement of repeated, in-run alignment with the real economy: \citet{wiesinger2010reverse} reverse-engineer heterogeneous minority-game traders against rolling windows of real Nasdaq data, with parameters and strategy mixtures re-optimized on each window before the model generates the next out-of-sample forecast; \citet{evans2025adage} formalize adaptive ABM as a Stackelberg game in which an outer environment-level optimizer repeatedly compares simulated outputs against experimental data and updates latent parameters to close the gap. The fifteen-year gap between them, and the very small number of qualifying works, is itself the central finding.

Even these two works, however, fall short of the level's full ambition. Empirical feedback in both cases enters the system at the model-fitting stage---rolling re-estimation of parameters, outer-layer recalibration of latent variables---rather than as a live signal that the running simulation continuously absorbs. A genuine Sim-to-Real Economic Twin would not stop at periodic re-fitting; it would let real-world observations enter the rollout itself, reshaping ongoing agent behavior, beliefs, and institutional dynamics as the world unfolds. By the stricter standard of live empirical feedback, the literature has yet to produce a single fully realized example, and the gap between current work and an authentic economic twin remains the most consequential frontier in the hierarchy.

\end{document}